\documentclass{article}

\usepackage[final]{corl_2026} 

\usepackage[utf8]{inputenc} 
\usepackage[T1]{fontenc}    
\usepackage{hyperref}       
\usepackage{url}            
\usepackage{booktabs}       
\usepackage{amsfonts}       
\usepackage{nicefrac}       
\usepackage{float}        
\usepackage{microtype}    
\usepackage{placeins}
\usepackage{graphicx}
\usepackage{amsmath}
\usepackage{amsthm}
\usepackage{amssymb}
\usepackage{mathtools}
\usepackage{enumitem}
\usepackage{xcolor}
\usepackage{wrapfig}
\usepackage{tikz}
\usepackage{float}
\usepackage[table]{xcolor}
\usepackage[dvipsnames]{xcolor}
\usepackage{makecell}
\usepackage{bm}

\hypersetup{
    colorlinks=true,
    citecolor=[rgb]{0.3,0.3,0.3},
    linkcolor=[rgb]{0.3,0.3,0.3},
    urlcolor=Blue,
}

\definecolor{CayenneClassic}{HTML}{DA474E}   
\definecolor{CayenneOrange}{HTML}{F0560D}    
\definecolor{CayenneSpicy}{RGB}{141, 63, 43} 
\definecolor{smartorange}{HTML}{DA7756}
\definecolor{smartpurple}{HTML}{800080}

\usepackage{tikz}
\usetikzlibrary{shapes,arrows}

\DeclareRobustCommand{\regselfplaymarker}{%
  \tikz[baseline=-0.6ex]{%
    \draw[solid, line width=1.3pt, color=CayenneClassic]
      (0,0) -- (0.55,0);%
    \fill[CayenneClassic] (0.275,0) circle (2pt);%
  }%
}

\DeclareRobustCommand{\regselfplaymarkerstripe}{%
  \tikz[baseline=-0.6ex]{%
    \draw[dashed, line width=1.5pt, color=CayenneClassic]
      (0,0) -- (0.55,0);%
  }%
}

\DeclareRobustCommand{\starmarker}{%
  \tikz[baseline=-0.6ex]{%
    \node[star, star points=5, star point ratio=2,
      fill=CayenneClassic, inner sep=1.6pt] at (0,0) {};%
  }%
}

\DeclareRobustCommand{\selfplaymarker}{%
  \tikz[baseline=-0.6ex]{%
    \draw[dashed, line width=1.5pt, color=black]
      (0,0) -- (0.55,0);%
  }%
}

\DeclareRobustCommand{\boxmarker}{%
  \tikz[baseline=-0.6ex]{%
    \fill[black] (-0.1,-0.1) rectangle (0.1,0.1);%
  }%
}

\DeclareRobustCommand{\smartmarker}{%
  \tikz[baseline=-0.6ex]{%
    \draw[solid, line width=1.3pt, color=smartpurple]
      (0,0) -- (0.55,0);%
    \fill[smartpurple] (0.275,0) circle (2pt);%
  }%
}

\DeclareRobustCommand{\smartpurplemarker2}{%
  \tikz[baseline=-0.6ex]{%
    \fill[smartpurple] (0,0) circle (3pt);%
  }%
}

\title{Human-like autonomy emerges from self-play \\ and a pinch of human data}

\author{
  Daphne Cornelisse$^{1}$ \And
  Julian Hunt$^{2}$ \And
  Zixu Zhang$^{3}$ \And
  Waël Doulazmi$^{4, 5}$ \And
  Kevin Joseph$^{2}$ \And
  Jaime Fern\'andez Fisac$^{3}$ \quad
  Eugene Vinitsky$^{1}$ \\ \\
  $^{1}$NYU Tandon School of Engineering \quad
  $^{2}$NYU Courant \quad
  $^{3}$Princeton University \\
  $^{4}$Centre for Robotics, Mines Paris \quad 
  $^{5}$Valeo
}

\begin{document}
\maketitle

\begin{abstract}
Self-play reinforcement learning has recently emerged as a way to train driving policies without any human data. It uses cheap, large-scale simulations to substitute expensive, large-scale human driving demonstrations. A key limitation of this approach is that policies trained through pure self-play can learn effective but alien driving conventions incompatible with people. Previous works attempt to mitigate such behavioral misalignments through extensive reward engineering and domain randomization, which are brittle and labor-intensive. Instead of completely discarding human demonstrations, our method treats them as a regularization objective on top of a minimal safe goal-reaching reward. Like the spice in a good stew, we find that a little human data goes a long way: our method uses only 30 minutes of human demonstrations, 2500× fewer than comparable imitation learning approaches. Resulting policies coordinate with held-out human trajectories and complete training in 15 hours on a single consumer-grade GPU. Videos and source code are available at  \url{https://spiced-self-play.com/}.
\end{abstract}


\section{Introduction}

\begin{wrapfigure}{r}[10pt]{0.44\textwidth}
    \centering
    \includegraphics[width=1\linewidth]{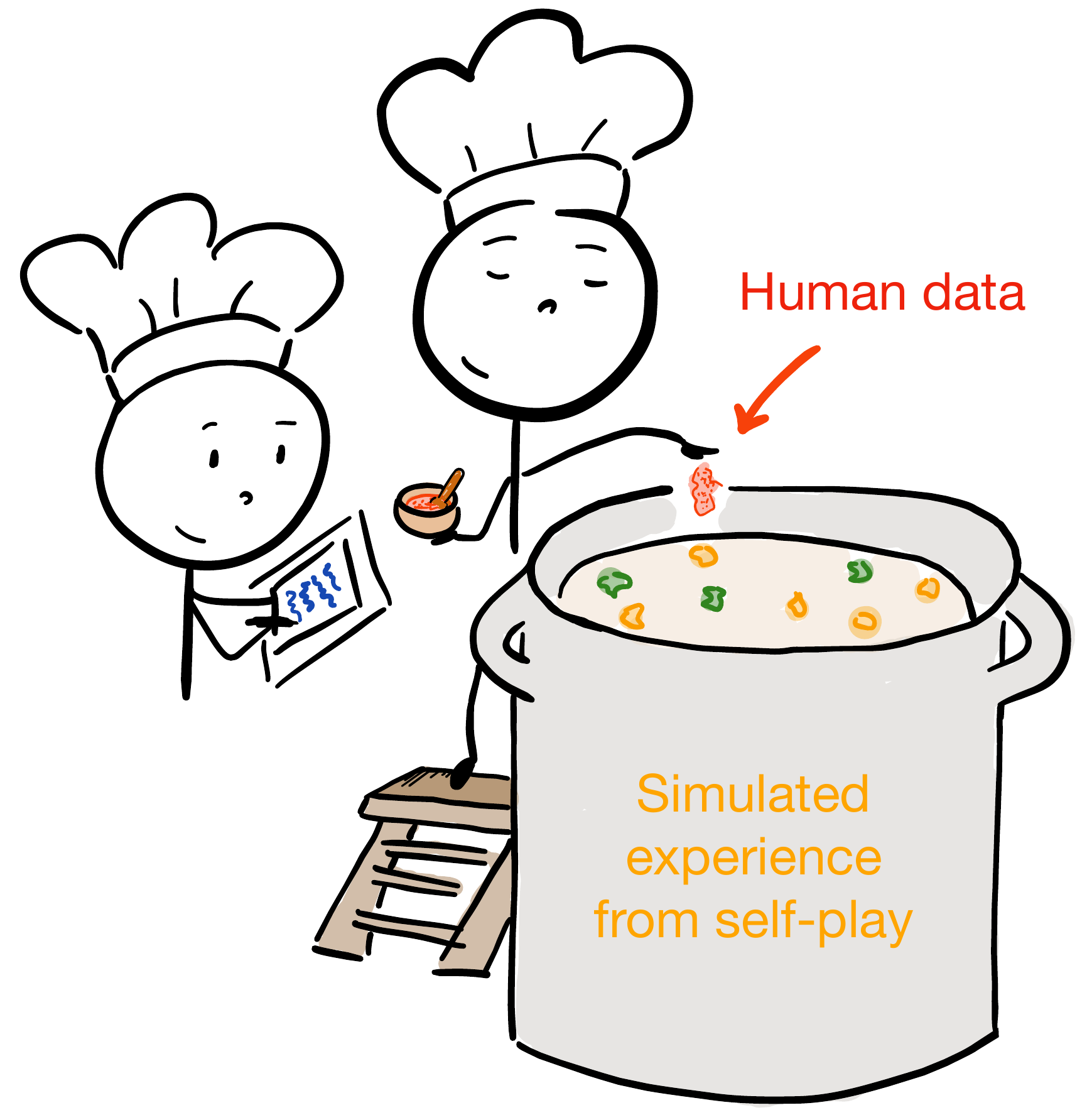}
    \label{fig:cartoon}
\end{wrapfigure}

Self-play reinforcement learning (RL) has produced superhuman agents in strategic games \citep{silver2017mastering, berner2019dota, DBLP:journals/corr/abs-2511-07312} and, more recently, has shown promise in real-world domains, such as autonomous driving \citep{kazemkhani2024gpudrive, cusumano2025robust, cornelisse2025building, chang2025spacer}. The approach elegantly sidesteps a central difficulty in multi-agent learning - how to model the opponent - through the following idea: the \textit{agent's opponent is a copy of itself}. The appeal here is that as the agent improves, so does its co-player. This gives rise to an automatically evolving curriculum \citep{leibo2019autocurricula} that takes the policy from random play to skilled behavior entirely through synthetic simulated experience.

In zero-sum games, this mechanism, with a sparse measure for success (e.g., +1 when winning a game of chess), is enough to produce strong play against arbitrary opponents. Many real-world settings, however, are not zero-sum. Driving, for instance, can be viewed as a mixed-motive game: each player has \textit{individual objectives} (reaching a destination safely) but must also \textit{coordinate} with other road users by adhering to shared norms, expectations, and conventions. Self-play RL with only a high-level objective for success provides no guarantees of such alignment; policies may converge to effective but ``alien'' strategies that are incompatible with human partners \citep{bakhtin2021no}. Concretely, an agent trained to ``reach a destination safely'' may very well learn to do so in reverse, sideways, or on the wrong side of the road if such constraints are not specified in the reward.

Previous works have addressed such misalignment in two ways. One line of work involves \textit{manual reward engineering}, where reward terms are added iteratively until the desired behavior and conventions emerge \citep{cusumano2025robust, qiu2026phase}. While effective, this strategy is labor-intensive by nature, domain-specific, and brittle since it is not trivial to figure out what reward will produce the desired human-like behavior \citep{knox2023reward}. A case in point is GIGAFLOW \citep{cusumano2025robust}, which required nine individually tuned reward terms and several other domain randomization techniques to produce naturalistic and cautious driving policies. On the other side of the spectrum, we have \textit{Imitation Learning} \cite{pomerleau1988alvinn, bojarski2016end, philion2023trajeglish}. In IL, the policy is optimized to \textit{directly imitate} human driving data, avoiding the need for defining a reward function altogether. However, robustness requires wide state coverage, so these approaches typically need large quantities of human demonstrations \citep{baniodeh2025scalinglaws}.

We take a different approach, grounded in a practical observation about the changing cost structure of experience generation. Modern RL frameworks and simulation infrastructure can generate between 300K and 20M environment steps \textit{per second} on a single consumer-grade GPU \citep{suarez2024pufferlib, pufferdrive2025github}, making synthetic experience generation effectively limitless. Human driving data, by contrast, requires manual collection and remains slow to scale. This suggests a natural role for human data in coordination games: not as the primary source of training signal, but as a lightweight anchor that steers the policy away from effective yet behaviorally alien strategies. Indeed, regularizing self-play RL toward such an anchor has shown promise in producing human-compatible agents in Diplomacy \citep{hu2022human, bakhtin2022diplomacy} and driving \citep{cornelisse2024hrppo, wang2026nomad, chang2025spacer}, yet \textit{how much} data is required to reach human compatibility remains, to our knowledge, unexamined.

\begin{figure}[ht]
    \centering
    \includegraphics[width=\linewidth]{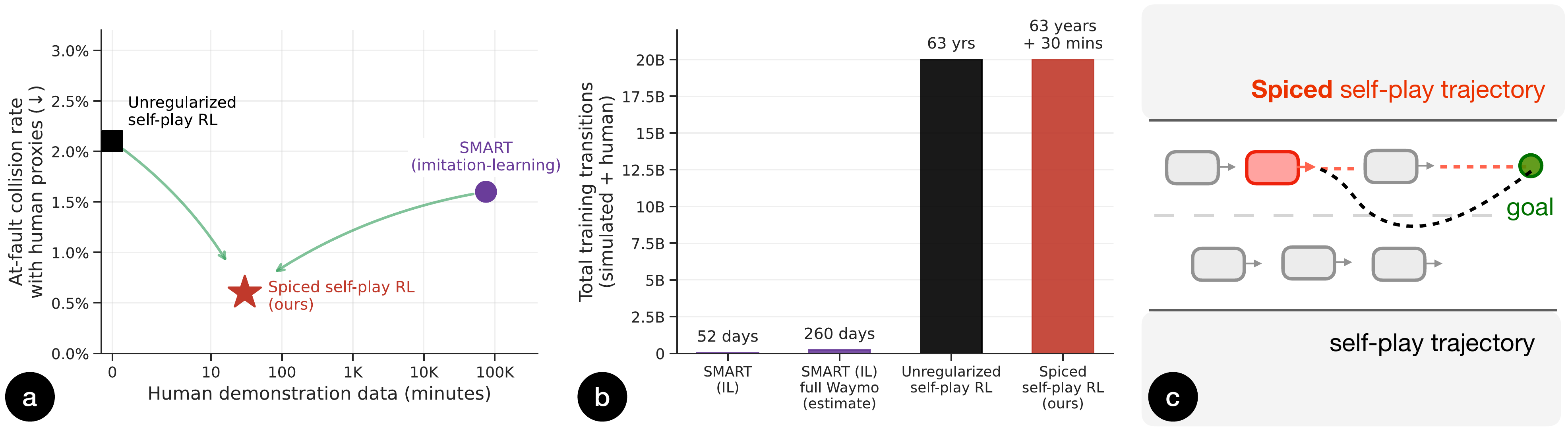}
    \vspace{-0.9em}
    \caption{\textbf{Spiced self-play RL policies achieve more human-like coordination with 30 minutes of human data and over 60 years of simulated experience.}
    (a) At-fault collision rate when interacting with human driving replays, as a function of the amount of human demonstration data used for training. Spiced self-play (\starmarker, ours) achieves an at-fault collision rate of $0.6\%$, compared with $2.1\%$ for unregularized self-play (\boxmarker) and $1.6\%$ for SMART-tiny CLSFT (\smartpurplemarker2) \citep{wu2024smart}, an imitation-learning approach trained on the full Waymo dataset plus a closed-loop fine-tuning step, but \textbf{not from multi-agent interactions}.
    (b) Total training data, including synthetic experience and human demonstrations. Both self-play variants use 20B simulated transitions (${\sim}63$ years at 10\,Hz); SMART uses 45M--225M human logged transitions (${\sim}52$ days--7 months; see Appendix~\ref{sec:exp_to_time_calc}).
    (c) Representative rollout (see \href{https://spiced-self-play.com/}{videos}). Resulting unregularized self-play policies (\selfplaymarker) may drive rather aggressively, threading through narrow gaps, while spiced self-play (\regselfplaymarkerstripe) waits for other agents to pass. The red vehicle is the controlled agent, conditioned on the green target destination; grey vehicles follow logged human trajectories.
    }
    \label{fig:hero_figure}
\end{figure}


We measure it. Anchoring self-play RL to human driving data from the Waymo Open Motion Dataset (WOMD) \citep{ettinger2021large}, we find that a surprisingly small amount of demonstration data improves coordination with human proxies. Paired with roughly 60 years of self-play experience, 30 minutes of human driving data (0.04\% of the full WOMD training set) yields a marked improvement, without doing any reward engineering or domain randomization. The effect mirrors an analogy already present in the literature: it is well documented that injecting a small fraction of detrimental data can cause catastrophic model degradation, a phenomenon known as \textit{data poisoning} \citep{wan2023poisoning, zhang2025persistent, souly2025poisoning}. To our knowledge, we are the first to report a comparable effect in the opposite direction within self-play RL; a small fraction of \textit{beneficial} data disproportionately improves behavior. Much like a pinch of cayenne changes the flavor of an entire dish, a small amount of human data appears to alter the behavior of a self-play policy. We refer to this as \textcolor{CayenneClassic}{\textit{data spicing}} and name our method \textcolor{CayenneClassic}{\textit{spiced self-play}}.

Concretely, we train a PPO policy \citep{schulman2017proximal} under a sparse reward for safe goal reaching, while \textit{regularizing} it toward a behavioral cloning anchor fit to a small amount of human driving data. We observe that:
\begin{itemize}[noitemsep, topsep=0pt]
    \item 30 minutes to 3 hours of human driving data, combined with self-play at scale, is sufficient to improve coordination with human proxies without reward engineering or domain randomization (Figure~\ref{fig:hero_figure}; Sections~\ref{sec:human_data_scaling_laws}, 
    \ref{sec:role_of_meta_data}).
    \item Spiced policies not only have lower collision rates, they also display more human-like behavior in terms of distributional realism~\citep{montali2023waymo} and collision severity profiles~\citep{waymo_safety_impact_2025} (Section~\ref{sec:full_bev_analysis}).
\end{itemize}

We open-source the codebase at \url{https://github.com/Emerge-Lab/spiced_self_play}. Policies can be trained end-to-end in 15 hours on a single consumer-class GPU.


\section{Related Work}

\paragraph{Imitation learning for autonomous driving.}
The generation of driving policies is a fundamental challenge across end-to-end autonomous driving \citep{chen2024end, jia2024bench2drive, hu2023planning, jiang2023vad}, multi-agent trajectory prediction \citep{huang2022survey}, and reactive traffic simulation \citep{montali2023waymo, vinitsky2022nocturne}. Driven by the widespread availability of large-scale human driving datasets \citep{caesar2019nuscenes, ettinger2021large, wilson2023argoverse}, imitation learning has become the dominant approach across all these domains \citep{bansal2018chauffeurnet}. Under this imitation learning paradigm, a broad spectrum of methodologies has emerged to fit models to historical data, ranging from marginal \citep{salzmann2020trajectron, gu2021densetnt, nayakanti2023wayformer} and joint \citep{ngiam2021scene, zhou2022hivt, shi2022motion, zhou2023query} forecasting to autoregressive sequence modeling of tokenized trajectories \citep{seff2023motionlm, philion2023trajeglish, wu2024smart} and continuous distribution learning via diffusion and promptable world models \citep{zhong2023guided, jiang2024scenediffuser, huang2024versatile, tan2025scenediffuser, liao2025diffusiondrive}. While these generative approaches yield diverse open-loop behaviors, they are fundamentally constrained by the scale of human data required and frequently suffer from compounding covariate shift in closed-loop deployment \citep{baniodeh2025scalinglaws}. To mitigate these shifts, recent hybrid approaches integrate reinforcement learning \citep{lu2023imitation, PengLLSGSF24, zhang2025closed}, yet they typically still rely on extensive human driving data as their primary optimization signal. Our approach systematically inverts this balance: rather than depending on human driving data as the core supervisor, we utilize synthetic, multi-agent RL self-play as the primary engine for discovering robust interactive behaviors, retaining a remarkably small human dataset strictly as a behavioral anchor to ensure conformity to realistic traffic norms.

\paragraph{Self-play reinforcement learning in games.}
Self-play reinforcement learning has produced superhuman agents in games from Go and Chess \citep{silver2017mastering, silver2018general} to Dota II \citep{berner2019dota} and Stratego \citep{DBLP:journals/corr/abs-2511-07312}, all without human data. Superhuman play is not the same as human-compatible play, however. Many games admit multiple equilibria, and self-play need not converge to those compatible with human partners \citep{bakhtin2021no, hu2020other}. This mismatch has been observed in cooperative games such as Hanabi \citep{bard2020hanabi}, as well as general-sum games such as Diplomacy \citep{bakhtin2021no} and simplified traffic navigation games \citep{DBLP:conf/aies/LererP19}. In these settings, self-play agents can develop internally consistent conventions that transfer poorly to interactions with human partners. The cause is reward underspecification: when the reward is defined as a score to maximize, there often exist different ways to achieve it. In other words, the solution space is large. Previous work attempts to resolve this by designing the reward by hand \citep{cusumano2025robust, qiu2026phase}. For instance, GIGAFLOW \citep{cusumano2025robust} demonstrates that reward engineering and domain randomization can produce naturalistic behavior at scale, at the cost of nine individually tuned reward terms. We avoid reward engineering entirely. A small amount of human data serves as a behavioral anchor, and self-play does the rest. This reduces a labor-intensive design problem to a one-hour data collection procedure.

\paragraph{Human-regularized self-play reinforcement learning and search.}
One alternative to reward engineering is to regularize self-play toward a human anchor policy. This idea has been explored in Diplomacy, where KL regularization toward a human prior during both search and learning produced agents that coordinate more effectively with human partners \citep{hu2022human, bakhtin2022diplomacy}. \citet{jacob2022modeling} study KL-regularized search more broadly and show that it recovers human-like play across several games. In autonomous driving, the idea has been explored at a limited scale \citep{cornelisse2024hrppo, chang2025spacer}. Previous work showed improved human-likeness and coordination with log-replays through regularized self-play RL in autonomous driving \citep{cornelisse2024hrppo}. However, the authors were bottlenecked by experience-generation speed: their simulator ran at 2,000 steps per second \citep{vinitsky2022nocturne}. As a result, the policies were trained on only 140 million self-play transitions across 200 scenarios, which required five days of wall-clock time and left little room to study data scaling. More recently, \citet{chang2025spacer} demonstrated that KL-regularized self-play can yield human-like driving policies using SMART \citep{wu2024smart} as the behavioral anchor. Notable differences to their setup include: 1) Vulnerable road users (VRUs; pedestrians and cyclists) were replayed from human data during training, which conflates the anchor's contribution with that of the mixed-in human trajectories and precludes a clean analysis of where the impact comes from; 2) Their behavioral anchor is a large tokenized model trained on the full 500,000-scenario Waymo dataset; 3) Policies were trained on 1 billion training transitions, particularly due to the high cost of running inference on SMART. We scale self-play to 20 billion steps, control all agents during self-play training to preclude human contamination of collected human data, and systematically study how much human anchor data is needed to improve human compatibility.

\vspace{-0.2cm}

\section{Method}

\vspace{-0.2cm}

\begin{figure}[ht]
    \centering
    \includegraphics[width=1\linewidth]{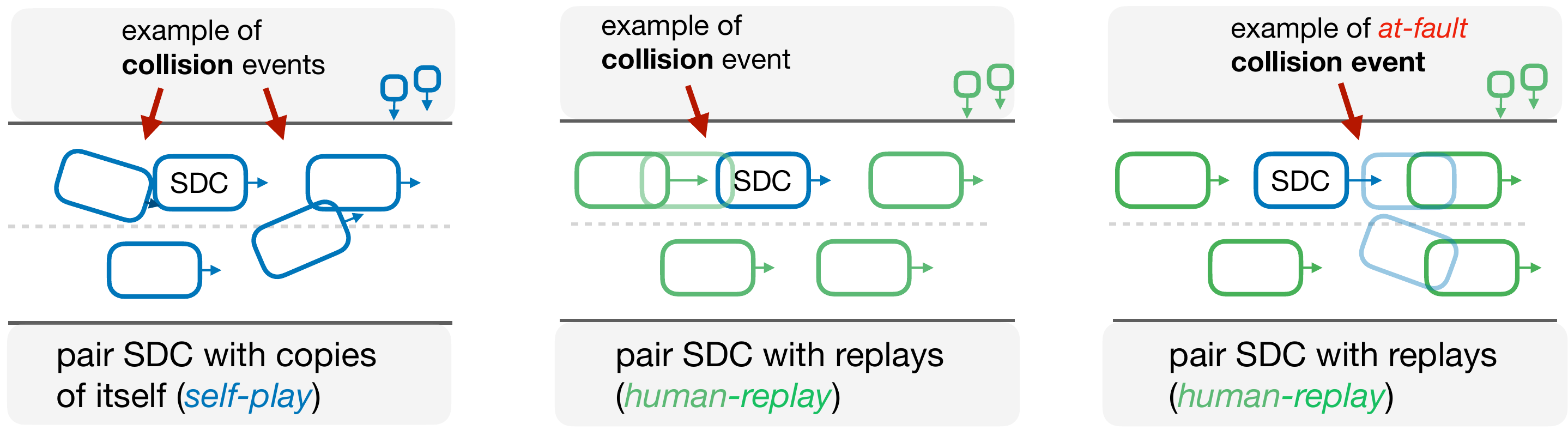}
    \caption[Self-play and human-replay evaluation settings.]{\textbf{Evaluation settings.} Self-play (left) and human-replay (center, right). Red arrows mark collisions. Rectangles are vehicles; squares are pedestrians. In human-replay, some collisions are effectively unavoidable: replay agents follow their logged trajectories and can drive into the controlled SDC from behind. We therefore distinguish between \textit{collisions} (any contact) and \textit{at-fault collisions} (contact caused by the controlled agent, following the NAVSIM benchmark~\citep{dauner2024navsim}).}
    \label{fig:cartoon_to_explain_evals}
\end{figure}

\paragraph{Problem setup.} A human-compatible agent should \textit{blend in} with human drivers. We approximate interaction with human road users by replaying logged human trajectories in simulation. We evaluate in three settings, illustrated in Figure~\ref{fig:cartoon_to_explain_evals}:
\begin{itemize}[noitemsep, topsep=0pt]
    \item \textbf{Self-play.} All agents are controlled by the \textbf{same policy} in a decentralized manner.
    \item \textbf{Human-replay.} Only the self-driving car (SDC) is controlled by the policy; all other agents follow their logged trajectories.
    \item \textbf{IDM.} The SDC is controlled by the policy; all other agents follow the Intelligent Driver Model~\citep{treiber_congested_2000}, following a precomputed lane-center path for lateral control and using longitudinal accelerations of IDM to maintain a safe gap between the lead vehicle \citep{charraut2025framework}.
\end{itemize}
An effective and human-compatible agent should reach its goal without collisions or off-road events across all three settings, each of which probes a distinct failure mode. Human-replay tests whether the policy has internalized human driving conventions against non-reactive co-players. IDM introduces closed-loop dynamics with reactive rule-based co-players. Self-play tests internal consistency and additionally serves as a convergence sanity check.

\paragraph{Metrics.} We report several metrics that capture task performance. The \textbf{score} is an aggregate metric; an agent scores 1 if it completes the task of driving to a goal destination before the end of the episode without colliding or going off-road, and 0 otherwise. To diagnose failure modes, we separately report \textbf{collision rate}, \textbf{at-fault collision rate}, \textbf{off-road rate}, and \textbf{route progress}. An ideal agent should score well with its own population as well as the human-replay population. Score-based metrics capture whether agents complete their task safely, but not whether their behavior looks human. We therefore also report \textbf{distributional realism} using the Waymo Open Sim Agent Challenge \citep{montali2023waymo} to compare their behavior to logged trajectories. Finally, we also analyze the severity of the at-fault collisions \cite{waymo_safety_impact_2025}. Metrics are reported on \textbf{held-out test scenarios} unless stated otherwise; see full definitions and details in Appendix~\ref{sec:eval_metrics}.

\subsection{Simulation Environment}

\paragraph{World initialization.} We use PufferDrive 2.0 \cite{pufferdrive2025github} for simulation and training. Environments are initialized from the Waymo Open Motion Dataset \citep[WOMD]{ettinger2021large}: each 9-second scenario provides a roadgraph, a variable set of agents (cars, cyclists, pedestrians) up to $N=32$, and per-agent initial poses and goals drawn from the logs. Each agent is goal-conditioned on a target destination ($x,y$ position) and receives a partial, decentralized, ego-frame observation consisting of its own state, the $N-1$ closest neighbors within 50\,m, and up to 128 nearby road segments (road edges, lanes and lines). World initialization and observation space details are provided in Appendix \ref{sec:world_init} and \ref{sec:observation_space}.

\paragraph{Reward function.}\label{sec:reward_function} To isolate the effect of human driving data, we avoid reward engineering and use a sparse reward: $+1$ for reaching the goal, $-1$ for collision or off-road events, and $0$ otherwise. Any differences in human-like behavior, therefore, stem from BC regularization rather than a hand-tuned reward. Episodes terminate once all agents reach their destinations, and we filter out transitions from agents that reach their goals early.

\subsection{Spiced Self-Play Reinforcement Learning}

\textcolor{CayenneClassic}{Spiced} self-play is \textit{regularized} self-play RL anchored to a small amount of human demonstration data (here driving logs). The anchor is a behavioral cloning policy fit to this data, which regularizes self-play through a KL penalty to the observation-conditioned anchor action probabilities. We train policies in two stages: a behavioral cloning (BC) anchor is first fit to human data, then frozen and used as a regularizer during self-play RL.

\paragraph{Step 1: Train the anchor policy.} To study how the amount of human data affects downstream performance, we train anchor policies on subsets of the full dataset $\mathcal{D} = \{(o_t^i, a_t^i)\}_{i=1}^{T \cdot K}$. We sample subsets $\mathcal{D}_n$ corresponding to $n$ scenarios, yielding roughly $\{10\text{ min}, 30\text{ min}, 3\text{ h}, 30\text{ h}\}$ of human driving data, and fit each anchor $\textcolor{smartpurple}{\tau_{\phi^n}}$ by minimizing negative log-likelihood:
\begin{align}
    \textcolor{smartpurple}{\phi^n} = \arg\min_{\phi} \sum_{(o^i_t,\, a^i_t)\, \in\, \mathcal{D}_n} -\log \tau_\phi(a_t^i \mid o^i_t).
\end{align}
We use only the self-driving car (SDC) trajectory from each scenario to generate our imitation data, as it is typically the highest-quality trajectory. Each anchor $\textcolor{smartpurple}{\tau_{\phi^n}}$ is then frozen for the subsequent self-play stage. Full details are in Appendix~\ref{sec:human_demonstration_collection}.

\paragraph{Step 2: Regularized self-play RL.} We train $\textcolor{CayenneClassic}{\pi_\theta}$ from scratch using Proximal Policy Optimization \citep[PPO]{schulman2017proximal}. The policy $\pi_{\theta}$ is represented by a 650k-parameter neural network. Each anchor $\textcolor{smartpurple}{\tau_{\phi^n}}$ serves as a behavioral regularizer via a KL penalty:
\begin{align}
    \mathcal{L}(\theta) = \mathcal{L}_{\mathrm{PPO}}(\theta) + \lambda \, \mathbb{E}_{o \sim \rho_{\textcolor{CayenneClassic}{\pi_\theta}}} \!\left[ D_{\mathrm{KL}}\!\left(\textcolor{smartpurple}{\tau_{\phi^n}}(\cdot \mid o) \,\Big\|\, \textcolor{CayenneClassic}{\pi_\theta}(\cdot \mid o)\right) \right],
\end{align}
where $\rho_{\textcolor{CayenneClassic}{\pi_\theta}}$ is the on-policy state distribution and $\lambda \geq 0$ controls regularization strength. The KL term pulls $\textcolor{CayenneClassic}{\pi_\theta}$ toward the anchor on states the policy actually visits, rather than on the offline distribution of $\mathcal{D}_n$. Hyperparameters and training details are in Appendices~\ref{sec:world_init} and~\ref{sec:appendix_training}.

\section{Experiments}
\vspace{-0.3cm}
This section summarizes the key results. Additional details and analyses are reported in the appendices. We structure the sections to answer the following questions:
\vspace{-0.4em}
\begin{enumerate}[noitemsep, topsep=0pt]
    \item \textit{Scaling human driving data for regularized self-play RL}: How much human data is needed for strong performance in both self-play and human-replay evaluations? (Section~\ref{sec:human_data_scaling_laws})
    \item \textit{Behavior and safety analysis}: How does a small amount of human demonstration data shape policy behavior beyond task performance? We analyze the effect on distributional realism, collision severity, and driving style (Section~\ref{sec:full_bev_analysis}).
    \item \textit{The role of metadata and scenario diversity}: Driving datasets such as WOMD and NuPlan provide \textit{scenario metadata}---road graphs and initial agent positions---that ground simulation at a fraction of the cost of collecting human driving data. How does the number of training scenarios (maps) used for self-play influence agent performance? (Section~\ref{sec:role_of_meta_data})
\end{enumerate}

\subsection{Scaling Human Driving Data for Regularized Self-Play RL}
\label{sec:human_data_scaling_laws}

How much collected human driving data does regularized self-play need, and how does this compare to imitation learning-only based approaches? It is worth noting that one reason the second question matters is that any apparent data efficiency on our side could simply reflect the homogeneity of the Waymo Open Dataset rather than an actual property of the method. We benchmark against unregularized self-play RL~(\selfplaymarker), a goal-conditioned RL policy that is trained to reach a goal without colliding with other agents or going off-road (Section \ref{sec:reward_function}). This provides a human-data-free lower bound. We also benchmark to SMART-tiny-CLSFT~\citep{wu2024smart, zhang2025closed}~(\smartmarker), the state-of-the-art IL approach in this domain. SMART is trained on the same nested driving data subsets; we additionally include the open-sourced SMART-tiny-CATK checkpoint~\citep{zhang2025closed}, trained on all 500k WOMD training scenarios, as an IL upper bound (Appendix~\ref{sec:smart_model_training_details}).

\begin{figure}[ht]
\centering
\includegraphics[width=1\linewidth]{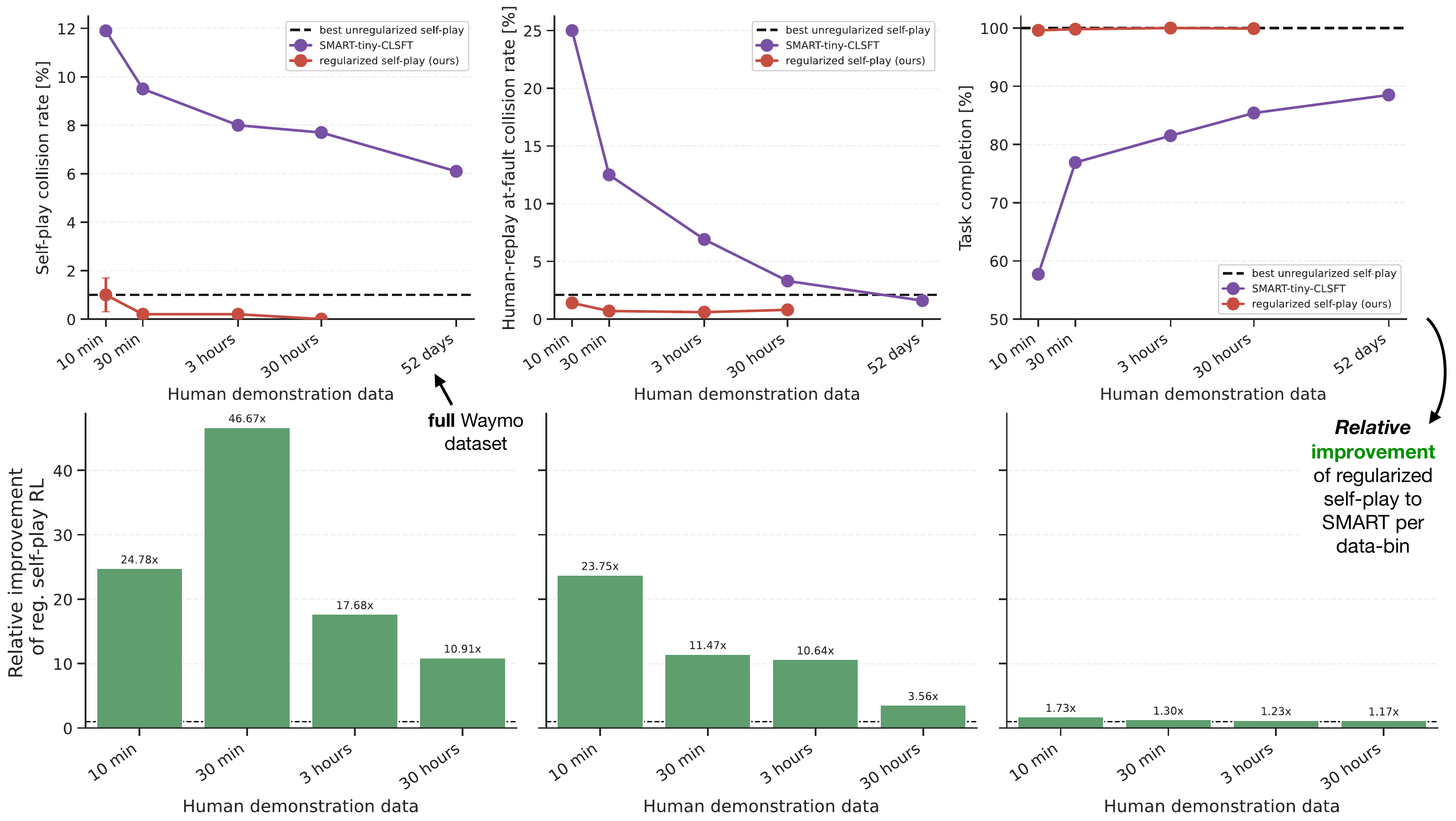}
\vspace{-0.9em}     
\caption{\textbf{Scaling human driving data for spiced self-play reinforcement learning.} \textit{Top}: Performance of \textit{Spiced} self-play RL~(\regselfplaymarker) and SMART with CAT-K closed-loop fine-tuning~(\smartmarker) as a function of total human log data used for training, evaluated in self-play and against human replays. Policies are evaluated on the same random 10k held-out WOMD validation split \citep{ettinger2021large}. Unregularized self-play RL~(\selfplaymarker) is shown as a horizontal line, since it uses no human driving data. The horizontal axis is semi-logarithmic. \textit{Bottom}: Relative improvement to IL baseline.}
\label{fig:human_data_requirements}
\end{figure}

\definecolor{tierbest}{HTML}{6FCF6A}
\definecolor{tiersecond}{HTML}{DFF04B}
\definecolor{tierthird}{HTML}{FBF4D0}
\definecolor{tierunregbest}{HTML}{D9D9D9}
\begin{table}[ht]
\centering
\caption{Performance versus amount of human demonstrations for the best trained policies on 10k held-out randomly sampled scenarios. For SMART, we report the best-performing variant at each data scale (details Appendix \ref{sec:smart_model_full_res}). Top-3 values per column are highlighted (\colorbox{tierbest}{best}, \colorbox{tiersecond}{2nd}, \colorbox{tierthird}{3rd}); the best value per column is additionally shown in bold. The unregularized self-play row uses no human driving data.}
\label{tab:human_data_results}
\vspace{-0.5em}  
\resizebox{\textwidth}{!}{%
\begin{tabular}{ll|rrr|rrrrr}
\toprule
 & & \multicolumn{3}{c|}{Self-play (test)} & \multicolumn{5}{c}{Human-replay (test)} \\
\makecell{Human demos \\ used} & Method & Coll. (\%) $\downarrow$ & Off-road (\%) $\downarrow$ & Route prog. (\%) $\uparrow$ & Score $\uparrow$ & Coll. (\%) $\downarrow$ & At-fault (\%) $\downarrow$ & Off-road (\%) $\downarrow$ & Route prog. (\%) $\uparrow$ \\
\midrule
10 min & SMART & 11.9 & 55.8 & 84.5 & 0.246 & 32.0 & 25.0 & 18.6 & 57.7 \\
30 min & SMART & 9.5 & 55.4 & 85.8 & 0.379 & 17.9 & 12.5 & 16.8 & 76.9 \\
3 hours & SMART & 8.0 & 53.6 & 86.2 & 0.518 & 11.4 & 6.9 & 4.5 & 81.5 \\
30 hours & SMART & 7.7 & 53.3 & 86.5 & 0.601 & 6.8 & 3.3 & 1.6 & 85.4 \\
52 days & SMART & 6.1 & 53.5 & 91.7 & 0.654 & 4.4 & 1.6 & \cellcolor{tiersecond} 1.1 & 88.5 \\
\midrule
--- & unreg. self-play & $1.0 \pm 0.4$ & \cellcolor{tierbest} $\bm{0.2 \pm 0.2}$ & \cellcolor{tierbest} $\bm{99.9 \pm 0.1}$ & $0.967 \pm 0.006$ & $2.7 \pm 0.5$ & $2.1 \pm 0.5$ & \cellcolor{tierbest} $\bm{0.6 \pm 0.2}$ & \cellcolor{tierbest} $\bm{100.0 \pm 0.0}$ \\
\midrule
10 min & reg. self-play (ours) & $1.0 \pm 0.7$ & \cellcolor{tierthird} $0.3 \pm 0.2$ & $99.0 \pm 0.4$ & $0.941 \pm 0.007$ & $3.9 \pm 0.6$ & $1.4 \pm 0.4$ & $1.4 \pm 0.4$ & $99.6 \pm 0.2$ \\
30 min & reg. self-play (ours) & \cellcolor{tiersecond} $0.2 \pm 0.1$ & $0.5 \pm 0.2$ & $99.3 \pm 0.3$ & \cellcolor{tierthird} $0.968 \pm 0.006$ & \cellcolor{tierthird} $2.0 \pm 0.4$ & \cellcolor{tiersecond} $0.7 \pm 0.3$ & $1.4 \pm 0.4$ & $99.8 \pm 0.1$ \\
3 hours & reg. self-play (ours) & \cellcolor{tierthird} $0.2 \pm 0.1$ & $0.6 \pm 0.4$ & \cellcolor{tierthird} $99.6 \pm 0.2$ & \cellcolor{tiersecond} $0.973 \pm 0.005$ & \cellcolor{tiersecond} $1.6 \pm 0.4$ & \cellcolor{tierbest} $\bm{0.6 \pm 0.2}$ & $1.2 \pm 0.3$ & \cellcolor{tiersecond} $100.0 \pm 0.0$ \\
30 hours & reg. self-play (ours) & \cellcolor{tierbest} $\bm{0.0 \pm 0.0}$ & \cellcolor{tiersecond} $0.3 \pm 0.2$ & \cellcolor{tiersecond} $99.7 \pm 0.2$ & \cellcolor{tierbest} $\bm{0.976 \pm 0.005}$ & \cellcolor{tierbest} $\bm{1.4 \pm 0.4}$ & \cellcolor{tierthird} $0.8 \pm 0.3$ & \cellcolor{tierthird} $1.1 \pm 0.3$ & \cellcolor{tierthird} $99.9 \pm 0.0$ \\
\bottomrule
\end{tabular}}
\vspace{-0.8em} 
\end{table}

\paragraph{Spiced self-play RL surpasses IL with a fraction of the human driving data.} As shown in Figure~\ref{fig:human_data_requirements} and Table~\ref{tab:human_data_results}, spiced self-play outperforms SMART-tiny-CLSFT \citep{zhang2025closed} across all data regimes and metrics. With as little as 30 minutes to 3 hours of human data, spiced self-play achieves the lowest at-fault collision rate (0.6-0.7\%); a 2.5$\times$ improvement over SMART-tiny-CLSFT trained on the entire Waymo train dataset (52 days; 1.6\%). The advantage is most pronounced at low human data: at 30 minutes, spiced self-play yields an 11$\times$ reduction in at-fault collision rate and 46$\times$ in self-play collision rate relative to SMART. Against standard self-play RL (at-fault CR: 2.1\%; \selfplaymarker), spiced self-play achieves a 3.5$\times$ improvement, demonstrating the value of an anchor trained on minimal human data as a regularizer. Regularized self-play RL with the 30-hour anchor leads to similar results.

\paragraph{Self-play exposes agents to a changing population of partners.}
The environment of a self-play RL policy is non-stationary: early policies have near-random behavior and become increasingly competent. This is in contrast to a single-agent RL setting, where the partner distribution is fixed. We observe that the self-play setting is associated with an increase in convergence to mutually consistent conventions. Spiced self-play agents achieve low collision rates in both self-play and cross-play with human logs (below 1.5\% in each). SMART, trained on 52 days of human data, incurs a 6\% self-play collision rate but only 1.6 \% when paired with logs. Two factors can explain this gap: sample count (20 billion transitions versus 225 million for SMART, Figure~\ref{fig:hero_figure}) and training paradigm (SMART is optimized open-loop for log-likelihood, then finetuned closed-loop to stay near the log distribution, and is never exposed to the partner distribution self-play naturally provides). To test for the role of the partner distribution, we compare self-play agents with agents trained \textit{directly against} the human-replay population (single-agent RL against static partners). The latter perform well within that population (at-fault collision rate 0.2--0.3\%) but do worse in self-play (0.8--1.2\%). This is consistent with exposure to reactive, evolving partners contributing to robustness (Figure~\ref{fig:single_multi_agent}).

\subsection{Behavior and Safety Analysis}
\label{sec:full_bev_analysis}

The goal of this section is to understand the behavioral differences between unregularized and regularized self-play policies beyond straightforward performance metrics.

\begin{figure}[htbp]
    \centering
    \includegraphics[width=1\linewidth]{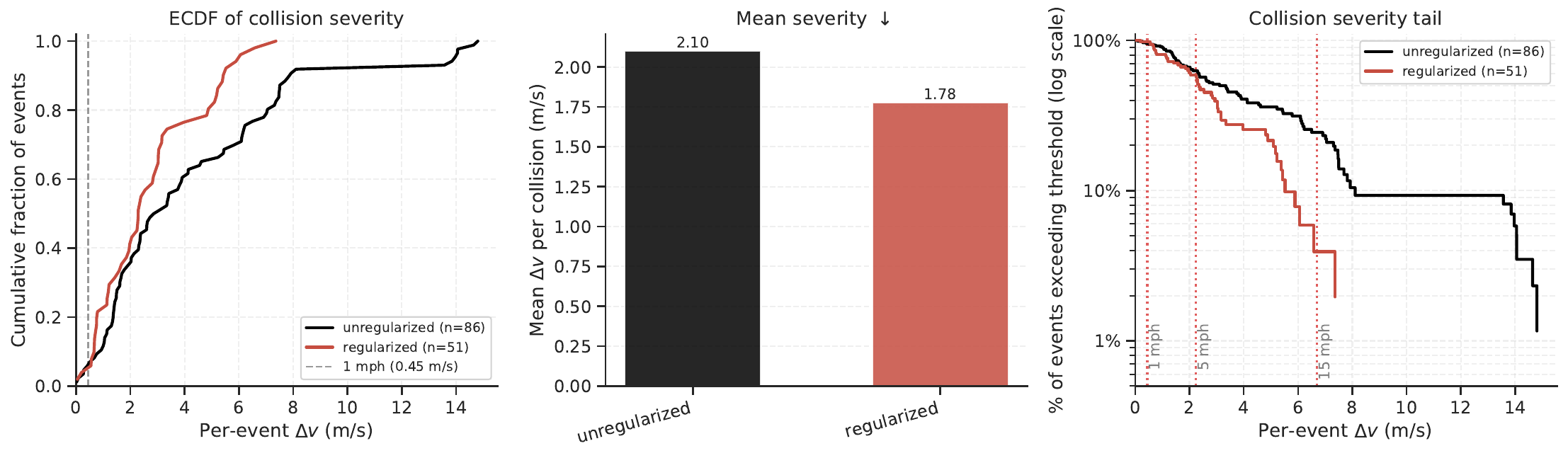}
    \vspace{-0.8em} 
    \caption{\textbf{Collision event severity.} \textit{Left:} empirical CDF of per-event $\Delta v$. The dashed line marks Waymo's $1$~mph ($0.45$~m/s) reporting threshold. \textit{Center:} mean $\Delta v$ per collision event, conditional on a collision occurring. \textit{Right:} fraction of collisions exceeding $\Delta v$ (log scale).}
    \label{fig:collision_severity_analysis}
    \vspace{-0.8em}     
\end{figure}

\paragraph{Spiced policies exhibit lower-severity collisions.} Collision rates, as reported in Sections \ref{sec:human_data_scaling_laws} and \ref{sec:role_of_meta_data}, measure how often agents fail, but not how bad the failures are. This distinction matters when policies are deployed alongside humans. Following Waymo's most recent safety report~\citep{waymo_safety_impact_2025}, we quantify collision severity via the \textit{change in velocity at impact} ($\Delta v$), a widely studied proxy for occupant injury risk. As shown in Table~\ref{tab:collision_severity} and Figure~\ref{fig:collision_severity_analysis}, regularization reduces both the frequency and the severity of failures. The mean per-event $\Delta v$ drops from $2.10$~m/s to $1.78$~m/s, and the maximum observed impact velocity falls from $14.80$~m/s to $7.36$~m/s. The improvement is more apparent when we focus on the tail of collision events: $23.3\%$ of unregularized collisions exceed $15$~mph, the threshold above which serious injury risk rises substantially, compared to $2.0\%$ for the regularized model. The survival curve in Figure~\ref{fig:collision_severity_analysis} (right) shows the two groups are nearly indistinguishable at low $\Delta v$, with the gap opening sharply above $5$~m/s and widening through the severe range. Regularization thus produces policies that not only collide less often but also cause less damage when they do collide.

\begin{table}[ht]
\centering
\caption{Collision severity tail breakdown on the random 10k WOMD validation split. This is the same dataset used in Table \ref{tab:human_data_results}. Best value is noted in \textbf{bold}.}
\label{tab:collision_severity}
\resizebox{\textwidth}{!}{%
\begin{tabular}{l|r|rrrrr}
\toprule
Method & \makecell{Events \\ (at-fault coll. rate)} & Mean $\Delta v$ (m/s) $\downarrow$ & Max $\Delta v$ (m/s) $\downarrow$ & $> 1$ mph (\%) $\downarrow$ & $> 5$ mph (\%) $\downarrow$ & $> 15$ mph (\%) $\downarrow$ \\
\midrule
unregularized & 213 (2.1\%) & 2.10 & 14.80 & \textbf{94.2} & 61.6 & 23.3 \\
regularized & 62 (0.6\%) & \textbf{1.78} & \textbf{7.36} & 96.1 & \textbf{56.9} & \textbf{2.0} \\
\bottomrule
\end{tabular}}
\end{table}

\paragraph{Regularized self-play improves realism with minimal data.} Unregularized self-play scores 0.680 on the WOSAC meta-score \citep{montali2023waymo}, with the largest deficits in the kinematic and interactive groups. Anchoring to 30 minutes of human data increases this to 0.725; the meta-score does not improve with additional data, suggesting BC anchor quality is the limiting factor. SMART-tiny CLSFT \cite{wu2024smart, zhang2025closed} achieves the highest realism score (0.755), yet underperforms on collision rate and task completion across every data bin (Section~\ref{sec:human_data_scaling_laws}), confirming that distributional similarity to logged human trajectories does not necessarily imply safety or competence \citep{cornelisse2025humanlikeness}. Additional results and graphs are in Appendix~\ref{sec:wosac_full}.

\paragraph{Regularized policies display more social driving behavior.}
We perform a qualitative analysis with representative videos available at \url{https://spiced-self-play.com/}. The most salient difference is that regularized policies are more considerate of surrounding traffic: they maintain greater following distances, avoid cutting in, and yield at intersections relative to unregularized self-play agents. RL policies are trained to maximize the expected cumulative \textit{discounted} return. An undesirable side-effect of this is that policies tend to achieve their task in the least number of steps possible. This is different than what humans do. A human driver will aim to get to her destination on time, but is not trying to get there as quickly as possible; \textit{satisficing} \citep{arumugam2024satisficing} rather than \textit{optimizing}. As visible in the videos and supported by the average episode length, regularization partially corrects for this: regularized agents complete their episodes in 64 steps on average ($\pm 3.5$), compared to 38 ($\pm 2.6$) steps for unregularized self-play.

This effect is also visible in the displacement errors to the human-replays in Table~\ref{tab:main_comparison}, which we decompose into a longitudinal component (along the direction of travel) and a lateral component (perpendicular to it). Lateral error reflects whether the policy follows the human's path through the scene (e.g., lane choice, turns) while longitudinal error reflects whether it travels that path at a human-like pace. A policy that rushes ahead stays on the right route but reaches each point too early or too late. We observe a clear difference: the unregularized longitudinal L2 (13.33 m) is over five times its lateral L2 (2.39 m). Regularization more than halves the longitudinal error (to 5.56 m) and nearly halves the lateral error (to 1.27 m), so the regularized policy follows human-like paths and traverses them at a human-like speed. The videos confirm both effects: the large longitudinal gap comes from unregularized RL policies driving very fast, and the lateral gap usually comes from their swerving around the replayed logs.

\section{Conclusion, Limitations \& Discussion}

\paragraph{Conclusion.} We consider a series of experiments aimed at putting the mixing of human driving data with synthetic simulated experience on a more scientific footing. Our central finding is that a small amount of human data, roughly 30 minutes to 3 hours of human driving data, can dramatically move the needle towards human-compatible driving agents. This is three orders of magnitude less than SOTA imitation learning baselines and is achieved without reward engineering or domain randomization techniques. The broader implication is that when simulation is cheap, and some clear metrics for desirable behavior are available, human driving data may be best used \textbf{not} as the primary training signal but as a \textit{lightweight anchor} that steers policies away from effective-but-alien equilibria.

\paragraph{Limitations.} 
\vspace{-0.3cm}
\begin{enumerate}[noitemsep]
    \item \textbf{Robustness in tight coordination scenarios}:  We perform an additional analysis to better understand the limitations of the resulting regularized policies. We curate a small dataset consisting of the top 200 most difficult interactive scenarios (see Appendix \ref{sec:eval_interactive_filtering}). Repeating the analysis from Section \ref{sec:human_data_scaling_laws} on this set of harder scenarios shows that, while the ranking of the policies holds (reg. self-play RL policies still outperform the SMART and unregularized self-play baselines by the same margins), the absolute at-fault collision rate increases from 0.7\% to 2.1-2.8\%. This indicates that there is room for improvement in the robustness of the resulting policies. Arguably, not all of these contacts reflect policy failures: some are caused by replay agents cutting abruptly into the SDC's lane, leaving almost no physically feasible avoidance response. What constitutes a fair collision-avoidance benchmark beyond at-fault heuristics is itself a difficult open question in both industry and academia~\citep{scanlon2026collision}. Nevertheless, an important direction for future work is to improve the robustness of regularized policies. See Appendix \ref{sec:extended_limitations} for the results, an in-depth discussion, and ideas to improve along this axis.
    \item \textbf{External validity of evals}: Our evaluations use human replays and IDM-controlled agents in simulation as proxies for coordination with humans. The extent to which performance in these settings transfers to on-road deployment remains an open question.
    \item \textbf{Sensitivity to the anchor}: Many underlying details by which regularizing the RL policy to the pre-trained BC anchor improves human-likeness remain incompletely understood. How do the properties of the anchor distribution, such as its entropy, affect the outcome? Results show that the regularized policies substantially outperform their anchors (see Figure \ref{fig:anchor_eval}, Table \ref{tab:anchor_results}), indicating that RL corrects for at least some suboptimal behavior in the anchor. It is unclear how sensitive this is to the BC policy's closed-loop quality, or how the correction occurs precisely.
\end{enumerate}

\paragraph{Combining human demonstrations with synthetic simulated experience.} Our key finding raises a deeper question that we have only touched the surface of, but is worth exploring further. Given the ability to generate simulated self-play experience on demand, what is the \textit{complementary value} of a bit of human data? Can we predict how much human data, and of what kind, is worth collecting for a given application X with structure Y? In the present work, we can loosely intuit two effects. First, the resulting regularized self-play RL policies are more human-like because the actor distributions stay close to the anchor distributions (see Section \ref{sec:solution_space}). Second, the resulting policies are more robust because they are exposed to broader coverage of the state space during training: the self-play agents learn from 20B transitions and start from random play, whereas the IL baseline is trained on a fixed dataset of 225 million expert transitions (Figure \ref{fig:hero_figure}, Center). But count is a crude explanation; not all transitions are equally informative. Recent work on epiplexity \citep{finzi2026entropy} takes a step toward formalizing this notion of data value, but in its current form, is a theoretical measure that we cannot yet compute or apply to data selection in practice. Developing tools to help determine \textit{what kind of human data} is needed to learn a given behavior, and \textit{predicting how much} is needed before collecting it, is a promising direction for future work.


\acknowledgments{We thank the authors of CAT-K~\citep{zhang2025closed} for generously sharing the weights of their best SMART-tiny-CLSFT checkpoint, which we use as the imitation learning baseline throughout the paper, and their code, which we use as a baseline for the scaling law experiments. We also thank Luke Rowe, Rodrigue de Schaetzen, and Roger Girgis for feedback on some early results and various interesting discussions on the topic of end-to-end driving and self-play. We thank Momchil Tomov for a helpful discussion on evals and metrics for evaluating human-likeness and compatibility in driving. Finally, we thank Zilin Wang and Nathan Monette for their helpful feedback on a draft of this work.

This work was also supported in part through the NYU IT High-Performance Computing resources, services, and staff expertise. Daphne Cornelisse is partially supported by the Cooperative AI Foundation and a Chishiki-AI SCIPE Fellowship.}

\bibliography{references}  

\newpage
\appendix

\section{Simulation Environment and Design}
\label{sec:appendix_simulation}

\subsection{World Initialization from Scenario Metadata}
\label{sec:world_init}
We use PufferDrive 2.0 for simulation and training \citep{pufferdrive2025github}. PufferDrive is a batched simulator that runs many environments in parallel, reaching 390k steps per second (SPS) on an NVIDIA RTX 5090 GPU. We initialize environments using the Waymo Open Motion Dataset (WOMD) \citep{ettinger2021large}, which provides a large set of multi-agent traffic scenarios. Each scenario supplies the metadata we need: the roadgraph, a variable number of agents (cars, cyclists, and pedestrians), and other objects in the scene. This information is the output of a perception stack, so we operate directly on these clean features (in bounding-box world).

Each scenario is 9 seconds long and discretized into 90 steps. We take each logged agent's initial position ($t=0$) as its starting position in the scene, and its last valid logged position ($t=T$) as its goal, which lets us goal-condition the agents. The full Waymo training dataset contains 500k scenarios, but in this paper we use at most 50k of the randomly sampled scenarios. When constructing the environments, we randomly sample scenarios from WOMD until we hit a target number of agents (e.g., on an NVIDIA RTX 4080 with 16GB of memory, we keep adding environments until we reach 1024 agents).

\subsection{Observation Space}
\label{sec:observation_space}
We take a decentralized approach and provide every agent with a partial view of the environment in a local coordinate frame. This is similar to the observation space of prior related works, such as GIGAFLOW \citep{cusumano2025robust}, and GPUDrive \citep{kazemkhani2024gpudrive}. At each timestep, an agent receives the combination of three feature blocks: an ego block describing its own state, a partner block describing the $N_p = 31$ closest other agents within a 50\,m radius, and a road block describing up to $N_r = 128$ nearby road segments drawn from a $21 \times 21$ grid of $5\,\text{m} \times 5\,\text{m}$ cells centered on the agent. Missing slots (fewer partners or road segments than the maximum) are zero-padded. Tables \ref{tab:obs_ego}, \ref{tab:obs_partner}, and \ref{tab:obs_road} list the features in each block. All positions and headings are expressed in the agent's local frame, so the observation is invariant to the global pose of the scene. The total observation vector has dimension $11 + 7 \times 31 + 7 \times 128 = 1{,}124$.

\begin{table}[ht]
\centering
\caption{Ego features (14 values) for the delta-local dynamics model. Features 0--3 expose the sampled conditioning variables to the policy so it can modulate its behavior as a function of $\lambda$ and the reward weights (Section~\ref{sec:role_of_meta_data}). We did not use conditioning in the paper and set all values to fixed values: $\lambda=0.075$; $r_\text{coll}, r_\text{off} = -1$ and $r_\text{goal} = +1$.}
\label{tab:obs_ego}
\begin{tabular}{rlll}
\toprule
Idx & Feature & Normalization & Description \\
\midrule
0     & $\lambda$                 & ---                & Human-regularization coefficient \\
1     & $r_\text{coll}$           & ---                & Sampled collision reward \\
2     & $r_\text{off}$            & ---                & Sampled off-road reward \\
3     & $r_\text{goal}$           & ---                & Sampled goal reward \\
4     & $\Delta x_\text{goal}$    & $\times 0.005$     & Goal position (ego frame), longitudinal \\
5     & $\Delta y_\text{goal}$    & $\times 0.005$     & Goal position (ego frame), lateral \\
6     & signed speed             & $/\,100\,\text{m/s}$ & Speed projected onto heading \\
7     & vehicle width            & $/\,15\,\text{m}$  & Ego bounding-box width \\
8     & vehicle length           & $/\,30\,\text{m}$  & Ego bounding-box length \\
9     & collision flag           & $\{0, 1\}$         & 1 if currently colliding \\
10    & entity type              & $/\,3$             & Vehicle (1), pedestrian (2), cyclist (3) \\
\bottomrule
\end{tabular}
\end{table}

\begin{table}[ht]
\centering
\caption{Partner features (7 values $\times$ 31 partners = 217 values). Partners are ordered by index and filtered to those within $50\,\text{m}$ of the ego agent. All positions and headings are in the ego frame.}
\label{tab:obs_partner}
\begin{tabular}{rlll}
\toprule
Idx & Feature & Normalization & Description \\
\midrule
0 & $\Delta x$                & $\times 0.02$      & Partner position, longitudinal \\
1 & $\Delta y$                & $\times 0.02$      & Partner position, lateral \\
2 & partner width             & $/\,15\,\text{m}$  & Partner bounding-box width \\
3 & partner length            & $/\,30\,\text{m}$  & Partner bounding-box length \\
4 & $\cos(\Delta\psi)$        & ---                & Relative heading, cosine component \\
5 & $\sin(\Delta\psi)$        & ---                & Relative heading, sine component \\
6 & partner signed speed      & $/\,100\,\text{m/s}$ & Signed speed along partner's heading \\
\bottomrule
\end{tabular}
\end{table}

\begin{table}[ht]
\centering
\caption{Road-segment features (7 values $\times$ 128 segments = 896 values). Segments are drawn from a $21 \times 21$ grid of $5\,\text{m}$ cells centered on the ego agent, and include road lanes, road lines, and road edges. Each segment is described by the midpoint, length, and orientation of a single polyline segment.}
\label{tab:obs_road}
\begin{tabular}{rlll}
\toprule
Idx & Feature & Normalization & Description \\
\midrule
0 & midpoint $x$             & $\times 0.02$      & Segment midpoint, longitudinal (ego frame) \\
1 & midpoint $y$             & $\times 0.02$      & Segment midpoint, lateral (ego frame) \\
2 & segment length           & $/\,100\,\text{m}$ & Length of the polyline segment \\
3 & segment width            & $/\,100\,\text{m}$ & Fixed nominal width (0.1\,m) \\
4 & $\cos(\theta)$           & ---                & Segment orientation in ego frame \\
5 & $\sin(\theta)$           & ---                & Segment orientation in ego frame \\
6 & segment type             & $\{0,1,2\}$        & Road lane (0), road line (1), road edge (2) \\
\bottomrule
\end{tabular}
\end{table}

\subsection{Actions and Dynamics}
\label{sec:action_space}

We use a single dynamics model with a discretized action space for both the unregularized and regularized agents.

\begin{figure}
    \centering
    \includegraphics[width=1\linewidth]{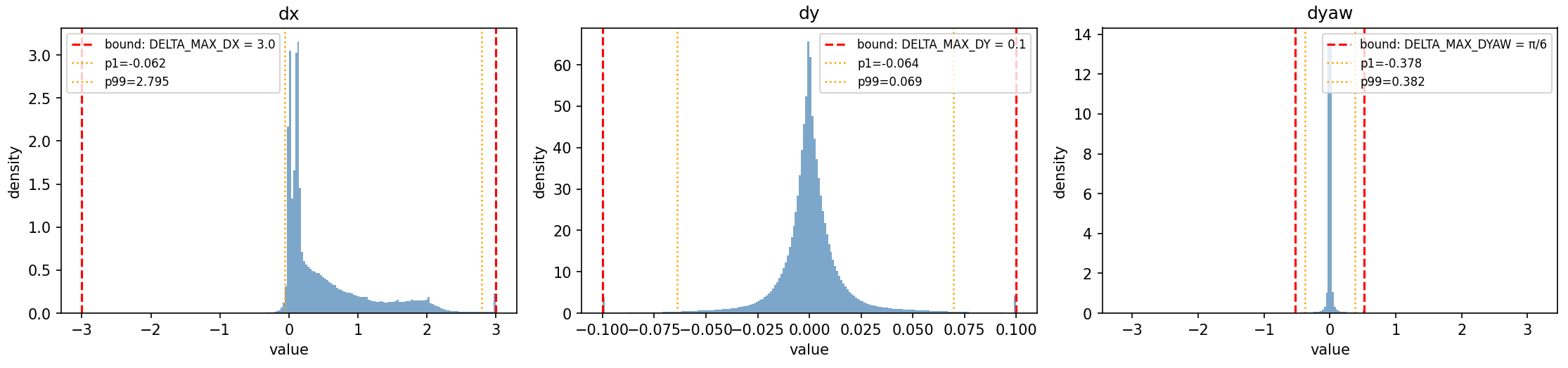}
    \caption{Discretized delta-local action space for each component ($\Delta x$, $\Delta y$, $\Delta \psi$). Histograms show the empirical density (blue) of 10{,}996{,}751 valid action timesteps recovered from expert trajectories across 10{,}000 maps. Yellow lines mark the 1st and 99th percentiles of the data; red lines mark the action-space bounds ($\pm 3.5$\,m, $\pm 0.1$\,m, $\pm \pi/6$\,rad). Each dimension is binned independently into 512 values. The bounds were chosen to respect natural movements in the data: $0.00\%$ of $\Delta x$ and $\Delta y$ samples fall outside their bounds, and $0.71\%$ of $\Delta \psi$ samples fall outside $\pm \pi/6$.}
    \label{fig:discretized_delta_local_action_space_bins}
\end{figure}

\paragraph{Delta-local dynamics with kinematic constraints.}
The action is a triple $(\Delta x, \Delta y, \Delta \psi)$ in the agent's local frame at time $t$. Translation is rotated into the world frame and added to the position; heading is updated directly:
\begin{align}
x_{t+1} &= x_t + \cos(\psi_t)\,\Delta x - \sin(\psi_t)\,\Delta y, \\
y_{t+1} &= y_t + \sin(\psi_t)\,\Delta x + \cos(\psi_t)\,\Delta y, \\
\psi_{t+1} &= \mathrm{wrap}(\psi_t + \Delta \psi).
\end{align}
Velocity is reported as the world-frame displacement divided by $\Delta t = 0.1$\,s. We bound each component roughly based on realistic actions present in the human data, as shown in Figure \ref{fig:discretized_delta_local_action_space_bins}; specifically, we define $\Delta x \in [-3.5, 3.5]$\,m, $\Delta y \in [-0.1, 0.1]$\,m, and $\Delta \psi \in [-\pi/6, \pi/6]$. Each of the three dimensions is binned independently into 51, 51, and 127 values, respectively. Figure \ref{fig:discretized_delta_local_action_space_bins} shows that the distributions for $\Delta y$ and $\Delta \psi$ are roughly symmetric, whereas the distribution for $\Delta x$ is strongly asymmetric. This is expected, since most vehicles move forward and only a small number of agents in the scenes drive in reverse (e.g., when parking).

Delta-local dynamics are kinematically unconstrained by default: the agent can translate laterally without rotating, pivot in place, or instantaneously reverse its heading rate. To prevent impossible behaviors, we apply two physics-based constraints to the action at each step. Each constraint clips the action after the previous one has been applied, with the previously executed (post-constraint) values used as the reference. The constraints are:

\begin{enumerate}
    \item \textbf{Longitudinal acceleration bound.} The change in implied forward speed is clipped to $\pm A_{\text{long,max}} \cdot \Delta t$, where $A_{\text{long,max}} = 8$\,m/s$^2$. This caps acceleration and braking.
    \item \textbf{Lateral motion envelope.} Lateral displacement is bounded by $|\Delta y| \le |\Delta x| \cdot \tan(\delta_{\max})$, where $\delta_{\max} = 0.7$\,rad is the maximum effective steering angle. This eliminates lateral sliding and side-shimmy at low forward speed.
\end{enumerate}

These physical constraints prevent kinematically implausible actions; they do not encode any preference over driving style and are independent of the human anchor.

\subsection{Collecting Human Driving Data}
\label{sec:human_demonstration_collection}

\begin{figure}[ht]
    \centering
    \includegraphics[width=1\linewidth]{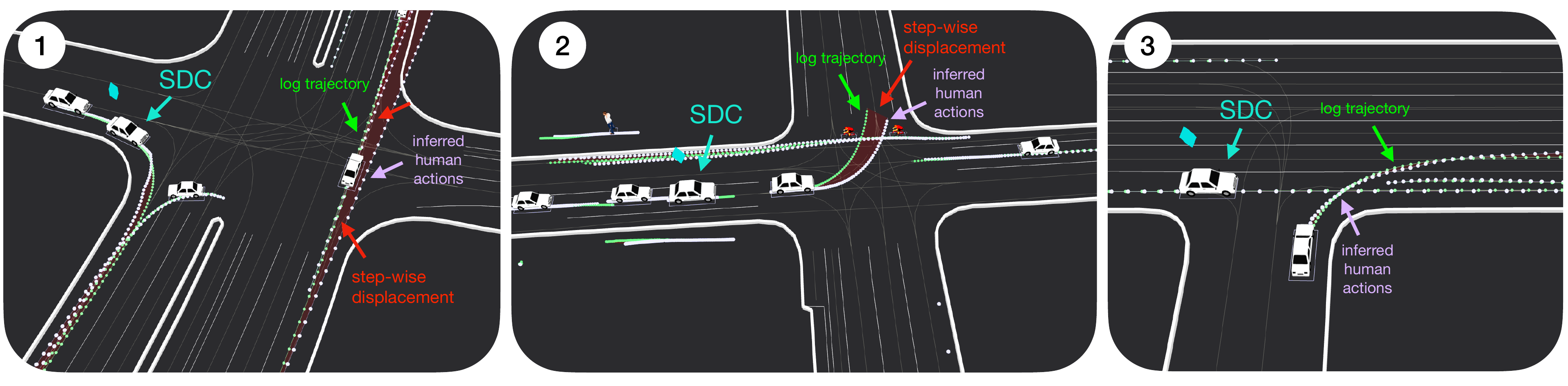}
    \caption{Three annotated example scenarios illustrating the human data collection process. The self-driving car (SDC), marked in cyan, is the Waymo vehicle whose human-driven trajectory we use as the driving log. Logged trajectories are shown in green; purple trajectories show the result of stepping each agent through the simulator under the inferred delta-local actions. We select only the SDC trajectory because it is typically the cleanest data in the scene; the visualized step-wise displacement illustrates a few low-quality (high-ADE) log trajectories that would otherwise contaminate the anchor.}
    \label{fig:anchor_data_quality_example}
\end{figure}

The behavioral cloning (BC) anchor is trained on observation--action pairs $(o_t, a_t)$. We therefore need actions that (i) live in the simulator's action space and (ii) reproduce the logged motion when applied through the simulator's dynamics. We construct the dataset in two steps. Figure \ref{fig:anchor_data_quality_example} shows three examples of this process in the simulator.

\paragraph{Step 1: Inferring actions from the data.}
For each timestep $t$, we invert the delta-local dynamics to recover the action that produced the next logged state. Projecting the world-frame displacement into the agent's local frame at $t$ gives:
\begin{align}
\Delta x_t &= \cos(\psi_t)(x_{t+1} - x_t) + \sin(\psi_t)(y_{t+1} - y_t), \\
\Delta y_t &= -\sin(\psi_t)(x_{t+1} - x_t) + \cos(\psi_t)(y_{t+1} - y_t), \\
\Delta \psi_t &= \mathrm{wrap}(\psi_{t+1} - \psi_t).
\end{align}
Each triple is clipped to the action bounds and snapped to the nearest discrete bin. Timesteps where either $t$ or $t+1$ is flagged invalid in the log are marked as missing and excluded from training.

\paragraph{Step 2: Replaying actions through the simulator.}
To produce observations, we replay the inferred action sequence through the simulator and record the observation at every resulting state.
The BC anchor is then trained on the resulting (simulator observation, inferred action) pairs. Discretization introduces a small error that grows inversely with bin size (details below); to prevent its accumulation, we instead \textit{teleport} agents to each ground truth successive state rather than stepping them forward with the inferred actions. We note that stepping agents directly is also viable when using larger action spaces, where the discretization error is smaller.

\begin{figure}[ht]
    \centering
    \includegraphics[width=1\linewidth]{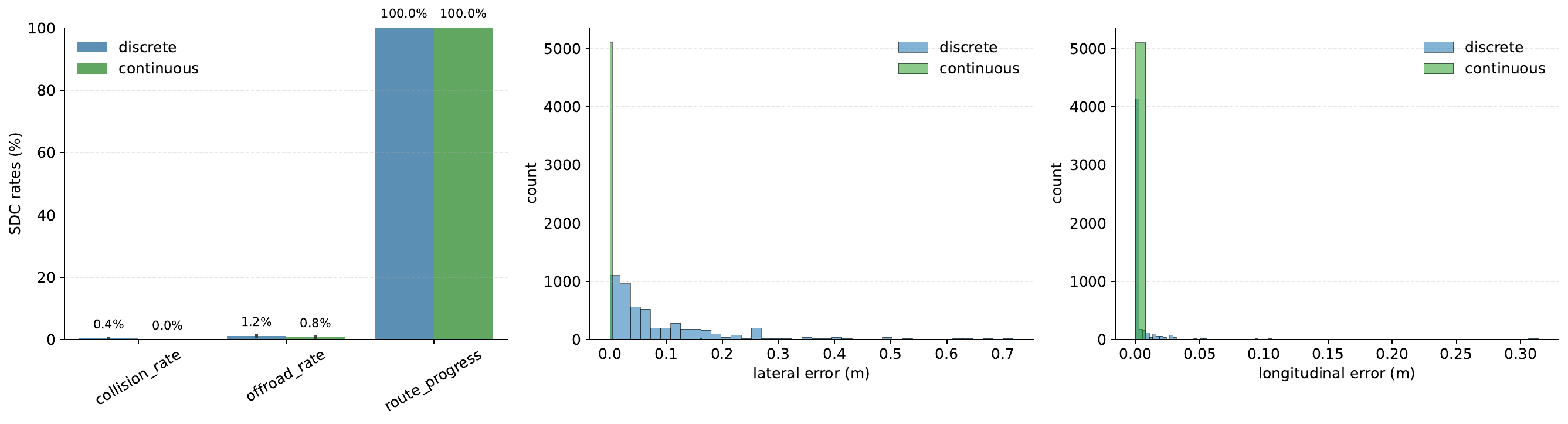}
    \caption{Effect of action discretization on inferred-expert-action quality. We replay each agent's logged trajectory through the simulator using actions inferred from the logs, comparing discrete (bin-quantized, blue) and continuous (direct float, green) action representations. \textbf{Left:} SDC rates aggregated across 10{,}240 pooled samples; both representations complete the route in 100\% of scenarios, but discretization induces modestly higher off-road and collision rates. \textbf{Center, right:} distributions of per-trajectory lateral and longitudinal L2 error to the logged pose. Continuous actions reproduce the log almost exactly (errors concentrated near zero), while discrete actions exhibit a small but consistent quantization floor of $\sim 0.1$\,m laterally. Error bars on the bar plot denote standard error.}
    \label{fig:effect_of_discretization_delta_local}
\end{figure}

\paragraph{Effect of discretization on performance.}
Figure~\ref{fig:effect_of_discretization_delta_local} and Table~\ref{tab:inferred_expert_actions} quantify the cost of discretization. Continuous actions reproduce the logged trajectory almost exactly (ADE $0.001$\,m), confirming that the delta-local dynamics and kinematic constraints are themselves well-posed. Discretizing into 512 bins per dimension introduces a quantization floor of ADE $0.097$\,m, which is roughly two orders of magnitude larger, but is still very close to the original trajectory. Off-road and collision rates increase modestly under discretization ($1.2\%$ vs.\ $0.8\%$ off-road, $0.4\%$ vs.\ $0.0\%$ collision), reflecting the rare cases where snapping to the nearest bin pushes the SDC just outside a road edge or into a static neighbor; both representations complete the route in $100\%$ of scenarios.

\begin{table}[ht]
\centering
\caption{Inferred-expert-action quality for the delta-local dynamics model. Comparison of discrete (bin-quantized) vs continuous (direct float) expert actions. Aggregated over 10,240 pooled samples. Values are mean $\pm$ SE.}
\label{tab:inferred_expert_actions}
\resizebox{\textwidth}{!}{%
\begin{tabular}{lrrrrrr}
\toprule
Action type & Route prog. (\%) $\uparrow$ & Coll. (\%) $\downarrow$ & Off-road (\%) $\downarrow$ & ADE (m) $\downarrow$ & Lateral L2 (m) $\downarrow$ & Longitudinal L2 (m) $\downarrow$ \\
\midrule
discrete & 100.0 & $0.4 \pm 0.1$ & $1.2 \pm 0.2$ & $0.097 \pm 0.002$ & $0.096 \pm 0.002$ & $0.004 \pm 0.000$ \\
continuous & 100.0 & 0.0 & $0.8 \pm 0.1$ & $0.001 \pm 0.000$ & $0.001 \pm 0.000$ & $0.001 \pm 0.000$ \\
\bottomrule
\end{tabular}}
\end{table}

\newpage
\subsection{Reward Function}

We use a sparse reward: $r^i = +1$ if agent $i$ reaches its goal within $\delta = 2$ meters before the episode ends, $-1$ on collision or going off-road, and $0$ otherwise. We deliberately omit dense shaping terms so that safe and human-compatible behaviors can emerge from regularization.

\section{Training}
\label{sec:appendix_training}

\subsection{Behavioral Cloning Anchor Policies}

Each anchor $\tau_n$ is trained by minimizing the negative log-likelihood of the logged actions under the factorized discrete action distribution described in Appendix~\ref{sec:action_space}. We extract observation, action tuples through the procedure described in Appendix~\ref{sec:human_demonstration_collection}. Note that we use only the SDC trajectory from each scenario for training, as it is the highest-quality data source. Since other agents are reconstructed from the perception stack, they exhibit more noise. Moreover, we have no guarantees about the driving quality of the surrounding humans. Since we obtain one trajectory per scene, each scenario contributes roughly 9 seconds of human data. Although these trajectories were collected in Waymo vehicles, they reflect manual human driving by an expert driver behind the wheel~\citep{ettinger2021large}.

\begin{figure}[ht]
    \centering
    \includegraphics[width=1\linewidth]{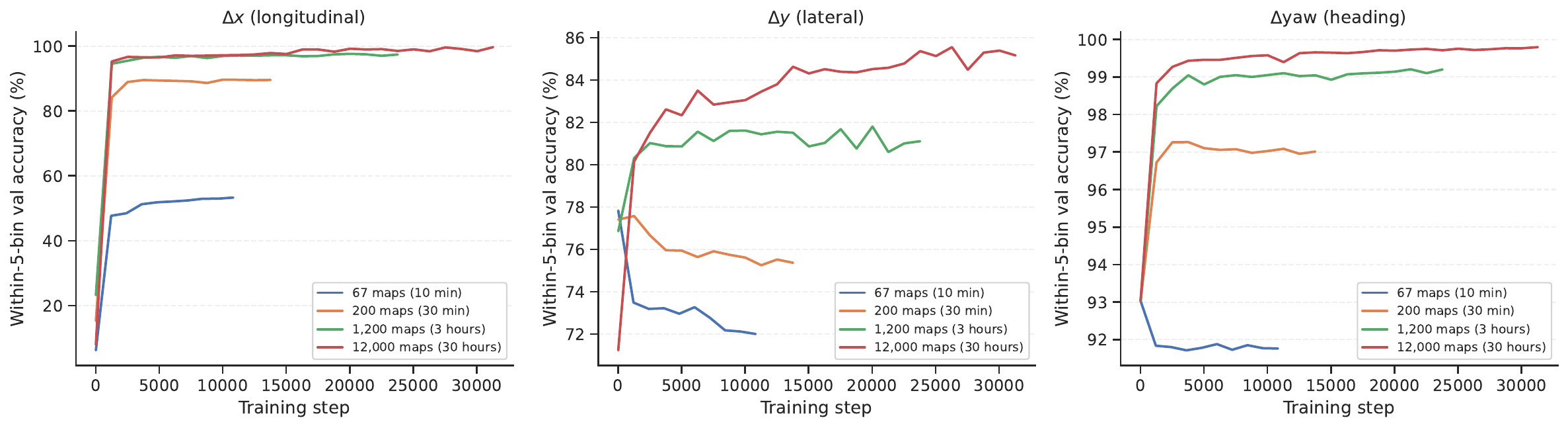}
    \caption{Training curves for the anchor policies. Each panel shows within-5-bin validation accuracy on a held-out set of scenarios for one action component ($\Delta x$, $\Delta y$, $\Delta \psi$). Curves terminate at different step counts because training stops once validation accuracy plateaus (no improvement for 100 consecutive epochs).}
    \label{fig:anchor_train_curves}
\end{figure}

We train with Adam at a learning rate of $10^{-4}$ and a batch size of 2048 for up to 5000 epochs, with early stopping on the held-out validation loss after 100 epochs without improvement. Table~\ref{tab:anchor_results} reports open- and closed-loop metrics for each anchor on 10{,}000 held-out validation scenarios. Figure~\ref{fig:anchor_train_curves} shows the 5-bin validation accuracy for each action head over training; from only 30 minutes of data, validation accuracy converges to between 80\% and 90\%. We use the 5-bin metric instead of top-1 as there are 256 bins per action head, so the step sizes between bins are very small.

\begin{figure}[ht]
    \centering
    \includegraphics[width=1\linewidth]{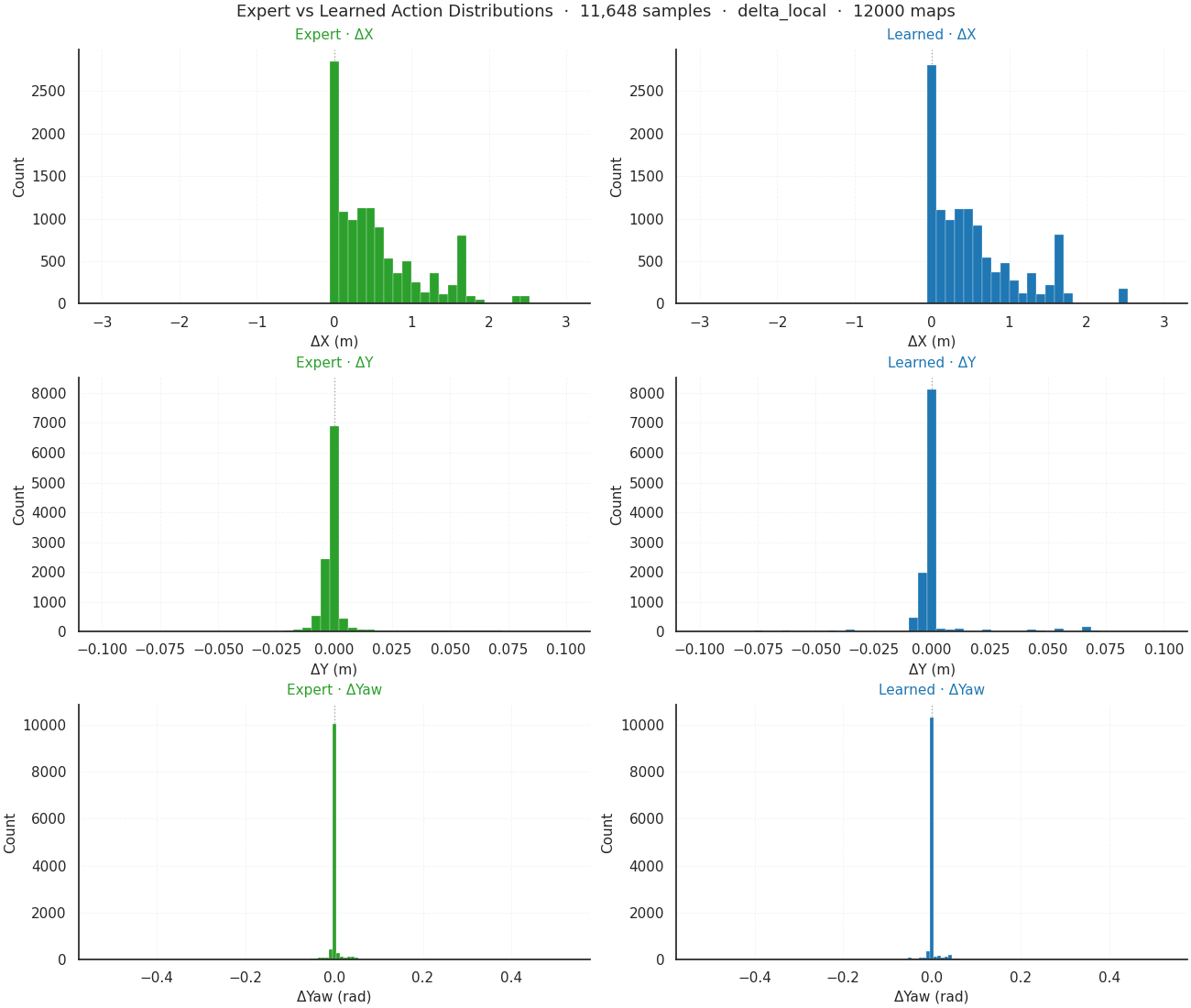}
    \caption{Example of actual vs. learned distributions - for 12k maps (30 hours)}
    \label{fig:learned_distributions_12k_maps}
\end{figure}

\begin{figure}[ht]
    \centering
    \includegraphics[width=1\linewidth]{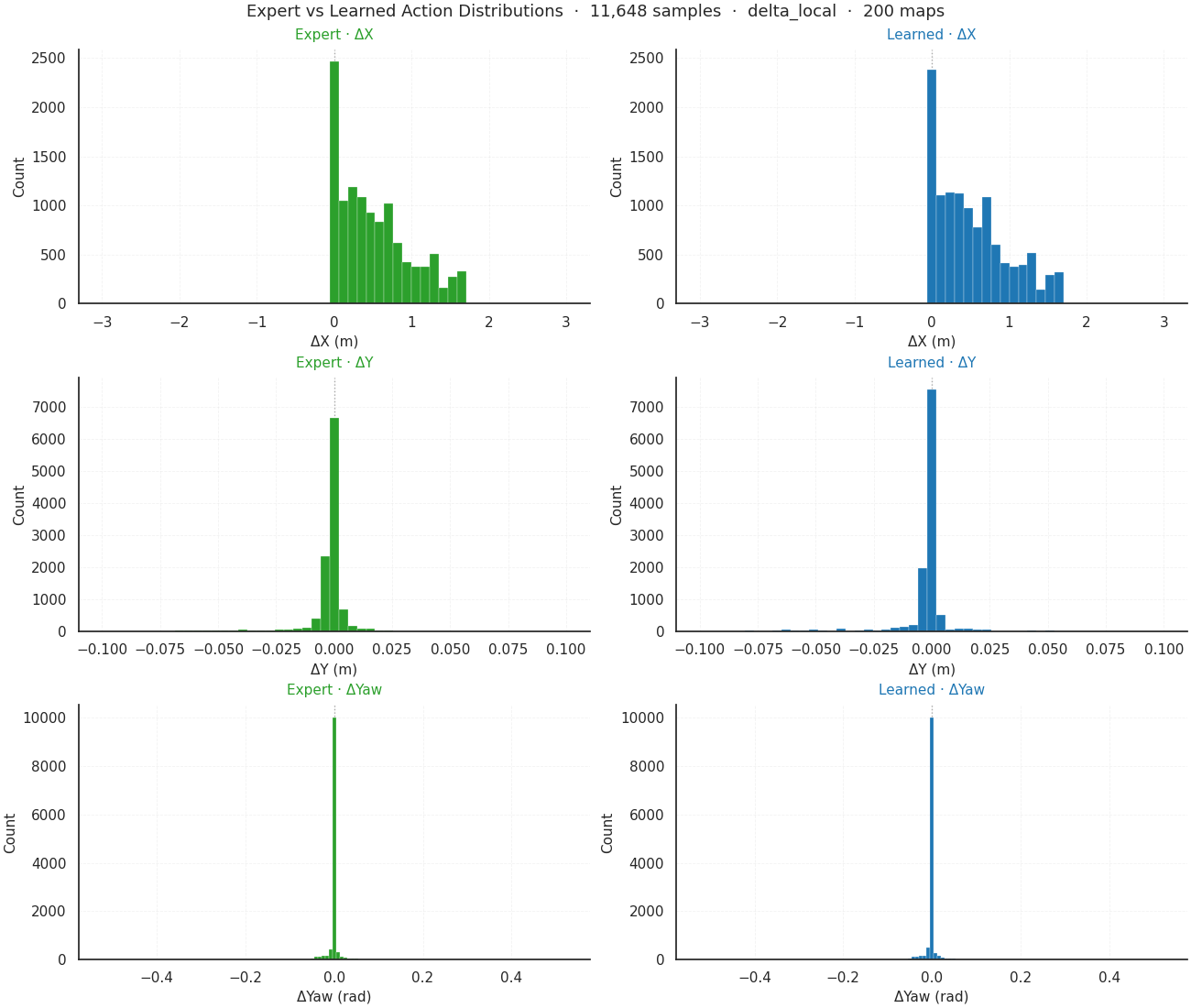}
    \caption{Example of actual vs. learned distributions - for 200 maps (30 min)}
    \label{fig:learned_distributions_200_maps}
\end{figure}

Figures~\ref{fig:learned_distributions_200_maps} and~\ref{fig:learned_distributions_12k_maps} compare the learned action distributions against the empirical distribution of the logged actions, for anchors trained on 30 minutes and 30 hours of data, respectively; in both cases, the learned distributions match the data reasonably well.

\begin{table}
\centering
\caption{BC anchor evaluation. Open-loop metrics on the held-out validation set; closed-loop metrics averaged over validation scenes. Within-5-bin accuracy is the average of $\Delta x$, $\Delta y$, $\Delta\mathrm{yaw}$ accuracies at the final training step.}
\label{tab:anchor_results}
\resizebox{\textwidth}{!}{%
\begin{tabular}{r|rrr|rr|rr}
\toprule
 & \multicolumn{3}{c|}{Open-loop} & \multicolumn{2}{c|}{Closed-loop self-play} & \multicolumn{2}{c}{Closed-loop human-replay (SDC only)} \\
Human data (h) & Acc. (\%) & Acc. $\pm 5$ bins (\%) & Loss & Route prog. & Score & Route prog. & Score \\
\midrule
0.2 & \cellcolor{green!5} 23.4 & \cellcolor{green!5} 72.4 & \cellcolor{red!50} 15.677 & \cellcolor{green!5} $0.720 \pm 0.012$ & \cellcolor{green!5} $0.215 \pm 0.013$ & \cellcolor{green!5} $0.765 \pm 0.007$ & \cellcolor{green!5} $0.242 \pm 0.009$ \\
0.5 & \cellcolor{green!24} 36.1 & \cellcolor{green!34} 87.3 & \cellcolor{red!17} 5.269 & \cellcolor{green!5} $0.719 \pm 0.011$ & \cellcolor{green!10} $0.277 \pm 0.014$ & \cellcolor{green!19} $0.800 \pm 0.006$ & \cellcolor{green!18} $0.371 \pm 0.011$ \\
3.0 & \cellcolor{green!42} 48.2 & \cellcolor{green!45} 92.6 & \cellcolor{red!6} 1.641 & \cellcolor{green!29} $0.835 \pm 0.010$ & \cellcolor{green!32} $0.502 \pm 0.017$ & \cellcolor{green!37} $0.842 \pm 0.006$ & \cellcolor{green!36} $0.538 \pm 0.011$ \\
30.0 & \cellcolor{green!50} \textbf{52.8} & \cellcolor{green!50} \textbf{94.9} & \cellcolor{red!5} \textbf{1.266} & \cellcolor{green!50} $\bm{0.932 \pm 0.007}$ & \cellcolor{green!50} $\bm{0.685 \pm 0.016}$ & \cellcolor{green!50} $\bm{0.873 \pm 0.006}$ & \cellcolor{green!50} $\bm{0.666 \pm 0.010}$ \\
\bottomrule
\end{tabular}}
\end{table}

\begin{figure}
    \centering
    \includegraphics[width=1\linewidth]{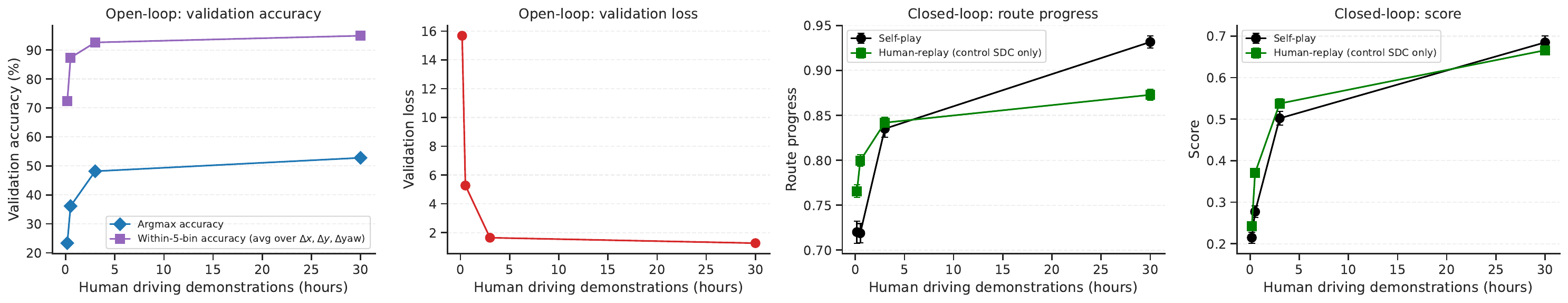}
    \caption{Open- and closed-loop performance of the anchor BC policies as a function of human driving data. Left: The final real (blue) and within 5 bin accuracy (purple) accuracy on 10,000 held-out validation scenarios. Right: Final validation loss. Right; Route progress; Right Score.}
    \label{fig:anchor_eval}
\end{figure}

\subsection{Self-Play Reinforcement Learning}
\label{sec:self_play_rl_training}

Both self-play variants run for 20 billion steps.

\subsubsection{Regularization}

Let $\pi_\theta$ denote the RL policy and $\tau_n$ the fixed BC anchor trained on $n$ scenarios. We regularize $\pi_\theta$ toward $\tau_n$ by adding a KL penalty on states visited during the rollout:
\begin{equation}
    \mathcal{L}_{\mathrm{reg}}(\theta)
    = \frac{\lambda}{M} \sum_{j=1}^{M}
      D_{\mathrm{KL}}\!\left(\tau_n(\cdot \mid o_j) \,\middle\|\, \pi_\theta(\cdot \mid o_j)\right),
\end{equation}
where $\lambda = 0.075$ is fixed throughout training and inference and $M$ is the minibatch size. The full objective augments standard PPO with this penalty:
\begin{equation}
    \mathcal{L}(\theta) = \mathcal{L}_{\mathrm{pg}} + c_v\,\mathcal{L}_{\mathrm{v}} - c_H\,H + \mathcal{L}_{\mathrm{reg}},
\end{equation}
where $\mathcal{L}_{\mathrm{pg}}$ is the clipped surrogate policy-gradient loss, $\mathcal{L}_{\mathrm{v}}$ the value-function loss, $H$ the entropy bonus, and $c_v$, $c_H$ their respective coefficients. The KL term pulls $\pi_\theta$ toward the anchor on states the policy actually visits, rather than on the offline logged data distribution. Setting $\lambda = 0$ recovers unregularized self-play.
\FloatBarrier
\subsubsection{Hyperparameters}
Table~\ref{tab:hyperparameters} lists the hyperparameters. We use the same parameters for regularized self-play RL and the baseline.

\begin{table}[ht]
\centering
\caption{PPO training hyperparameters.}
\label{tab:hyperparameters}
\resizebox{0.8\textwidth}{!}{%
\begin{tabular}{llllll}
\toprule
\multicolumn{2}{l}{\textit{Architecture}} & \multicolumn{2}{l}{\textit{Training}} & \multicolumn{2}{l}{\textit{Environment \& Rewards}} \\
\midrule
Input size          & 64          & Total timesteps        & 20B                   & Number of agents    & 1,024 \\
Hidden size         & 256         & Batch size             & 524,288               & Number of workers   & 16 \\
RNN type            & LSTM        & Minibatch size         & 32,768                & Episode length      & 150 steps \\
RNN input size      & 256         & Rollout horizon        & 32                    & Timestep $\Delta t$ & 0.1 s \\
RNN hidden size     & 256         & Update epochs          & 1                     & Goal radius         & 2.0 m \\
                    &             & Learning rate          & $4.26 \times 10^{-3}$ & Action space        & Discrete \\
                    &             & LR schedule            & Linear annealing      & Dynamics model      & Delta-local \\
                    &             & Adam $\beta_1$         & 0.9                   & Goal reward         & $+1.0$ \\
                    &             & Adam $\beta_2$         & 0.999                 & Collision penalty   & $-1.0$ \\
                    &             & Adam $\epsilon$        & $10^{-8}$             & Off-road penalty    & $-1.0$ \\
                    &             & Clip coefficient       & 0.2                   &                     & \\
                    &             & Entropy coefficient    & 0.001                 &                     & \\
                    &             & VF coefficient         & 2.0                   &                     & \\
                    &             & VF clip                & 0.2                   &                     & \\
                    &             & GAE $\lambda$          & 0.95                  &                     & \\
                    &             & Discount $\gamma$      & 0.99                 &                     & \\
                    &             & Max gradient norm      & 1.0                   &                     & \\
                    &             & Priority $\alpha$      & 0.85                  &                     & \\
                    &             & Priority $\beta_0$     & 0.85                  &                     & \\
                    &             & V-trace $c$ clip       & 1.0                   &                     & \\
                    &             & V-trace $\rho$ clip    & 1.0                   &                     & \\
                    &             & Optimizer              & Adam                  &                     & \\
                    &             & Seed                   & 42                    &                     & \\
\bottomrule
\end{tabular}}
\end{table}

\subsection{SMART Model Training and CATK finetuning}
\label{sec:smart_model_training_details}

\paragraph{IL data scaling baseline experiments.} We trained SMART models via CATK \citep{zhang2025closed} using the open-sourced codebase \url{https://github.com/NVlabs/catk} at commit \texttt{d23886761fc5b5628c5973148c40284452745745}. For the data scaling experiments, we used subsets of the Waymo Open Motion Dataset (WOMD). WOMD motion shards were preprocessed into CATK's per-scenario cached format, and all training subsets were constructed from these cached scenario files.

Our final local runs used the \texttt{smart\_mini\_3M} model with vehicle-only supervision on deterministic subsets of 67, 200, 1200, and 12000 scenarios. In the subset construction scripts, scenarios are sorted by cached scenario filename in lexicographic order before selecting subsets. Vehicle-only supervision means that only vehicle agents contribute to the training loss, while pedestrians and cyclists remain present in the scene and are available as contextual inputs to the model. The local models were trained with CATK's \texttt{pre\_bc} configuration on a single GPU for 64 epochs with batch size 8.  Results for both the SMART behavioral cloning checkpoints and the CAT-K / CLSFT fine-tuned checkpoints are reported in Table \ref{tab:smart_finetuning_results}.

\paragraph{Open-sourced checkpoints.} We additionally compare against two author-provided checkpoints: a behavioral cloning checkpoint (\texttt{pre\_bc\_E31.ckpt}) and a closed-loop supervised fine-tuning checkpoint (\texttt{clsft\_E9.ckpt}). For downstream evaluation, we exported predictions as \texttt{.pkl} files on the same 10k random validation split (data available at \url{https://huggingface.co/datasets/daphne-cornelisse/pufferdrive_womd_val}). We use two export modes: an all-agents mode, where the model controls all agents, and a planning mode, where only the SDC is controlled by the model while all other agents are replayed from ground truth. We re-exported all combinations of models and export modes with 32 rollouts for multimodal evaluation. We verified that there is zero scenario-ID overlap between each local training subset (67, 200, 1200, and 12000 scenarios) and the evaluation set.

\section{Neural Network Architecture}
\label{sec:appendix_architecture}

Both the BC anchor and the RL policy share the same multi-modal encoder structure. The flattened environment observation vector is first unpacked into its modalities: ego state, partner agents, and road segments. Each modality is processed by a two-layer MLP with ReLU activation and layer normalisation between the two linear layers. Partner and road embeddings are then aggregated across objects via max-pooling, producing one vector per stream. The three pooled vectors are concatenated and passed through a shared two-layer MLP (Linear $\to$ ReLU $\to$ Linear) to produce the final embedding. Separate linear heads decode this embedding into logits over each action dimension; a separate linear head with unit output produces the value estimate. The two architectures differ in width and in the presence of recurrence:

\begin{itemize}
    \item \textbf{BC anchor.} Per-stream MLP width 128, shared MLP $3{\times}128 \to 512 \to 512$. No recurrence. Actor heads are linear projections from the 512-dimensional embedding. It has 776,190 trainable parameters.
    \item \textbf{RL policy.} Per-stream MLP width 64, shared MLP $3{\times}64 \to 256 \to 256$. The 256-dimensional embedding is passed through a single-layer LSTM with input size 256 and hidden size 256 (PufferLib \texttt{LSTMWrapper}). Actor and critic heads are linear projections from the 256-dimensional LSTM output. It has 650k trainable parameters.
\end{itemize}

Road segment features include a categorical type field that is replaced by a 7-class one-hot vector before encoding, expanding the road feature dimension from $d_\text{road}$ to $d_\text{road} + 6$.

\section{Evaluation}

\subsection{Filtering the Waymo Dataset for Interactive SDC Scenarios}
\label{sec:eval_interactive_filtering}

As pointed out in earlier works \cite{distelzweig2026beyond, cornelisse2024hrppo}, many scenarios in the Waymo Open Motion Dataset (WOMD) involve the self-driving car (SDC) traveling without meaningful interaction with other agents---the SDC reaches its destination without requiring coordination or yielding. To increase the signal in our human-replay evaluation, we filter the dataset for scenarios in which the SDC trajectory intersects with other agents' trajectories, indicating situations that require coordination, such as merging, yielding, or navigating busy intersections.

\begin{figure}
    \centering
    \includegraphics[width=\linewidth]{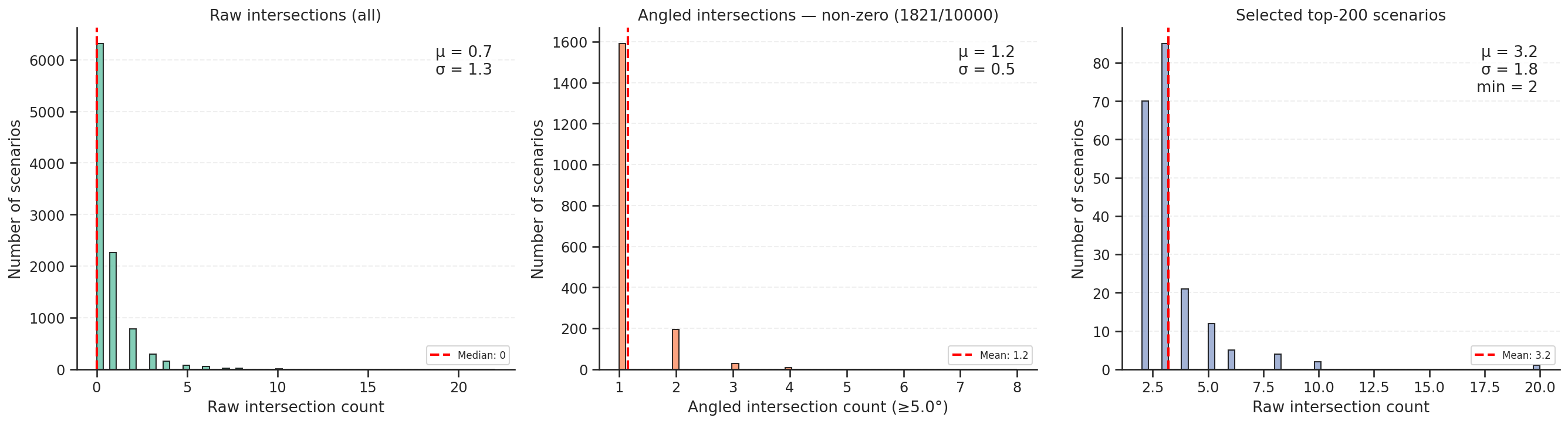}
    \caption{Distribution of SDC trajectory intersection counts. \textbf{Left:}~raw intersection counts across all 50k scenarios. \textbf{Center:}~angled intersections (non-zero only). \textbf{Right:}~distribution within the selected top-200 subset.}
    \label{fig:interactivity_distribution}
\end{figure}

\begin{figure}
    \centering
    \includegraphics[width=0.32\linewidth]{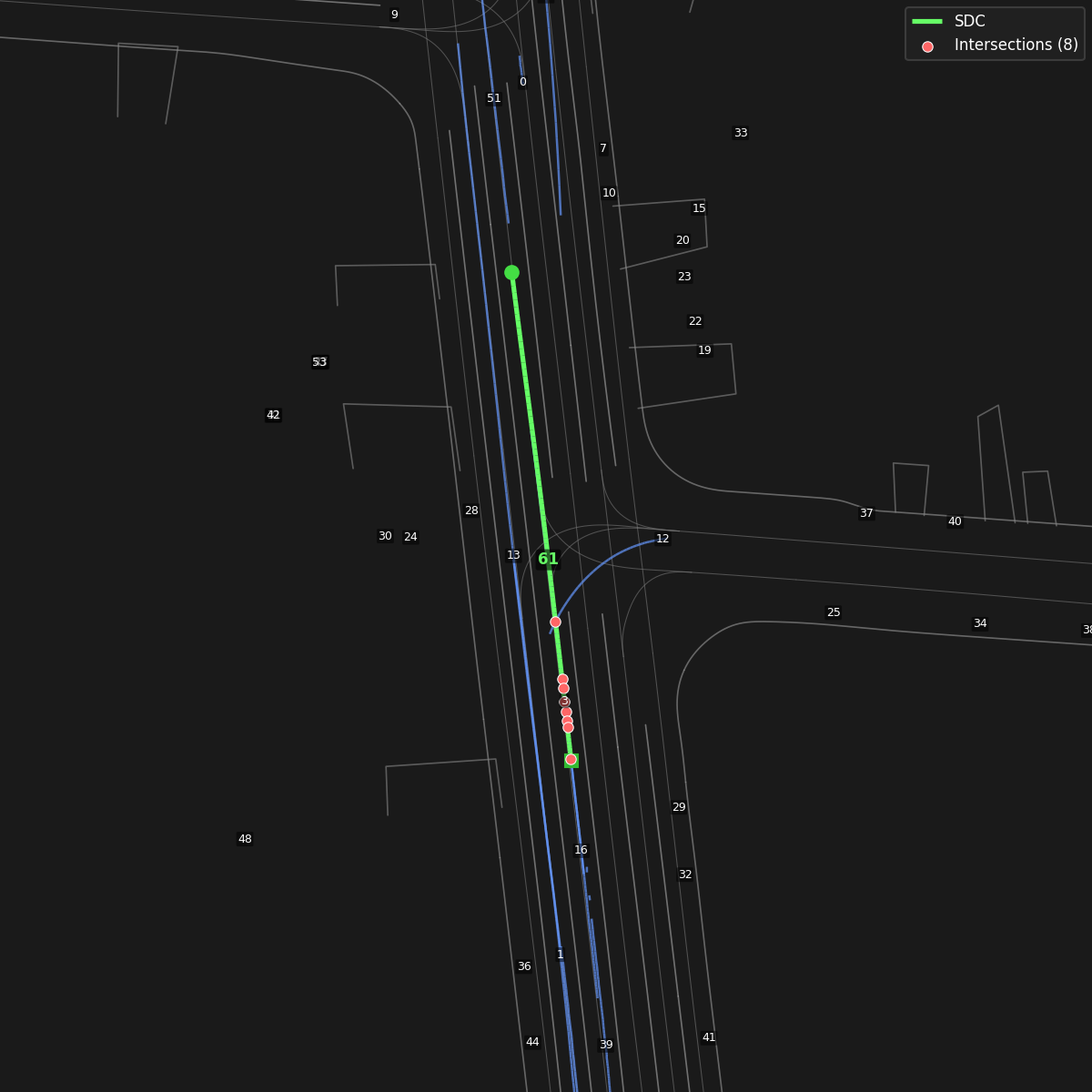}%
    \hfill
    \includegraphics[width=0.32\linewidth]{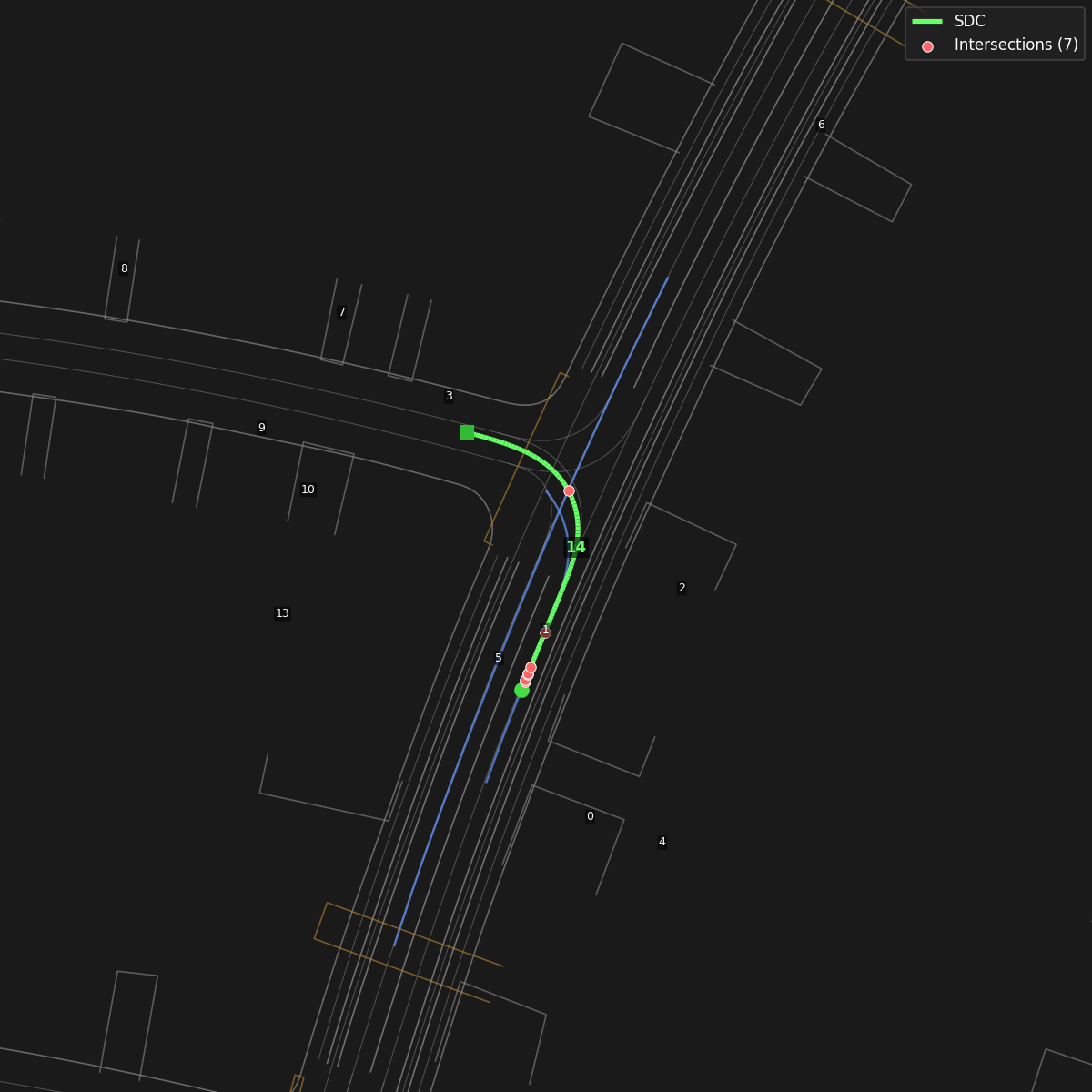}%
    \hfill
    \includegraphics[width=0.32\linewidth]{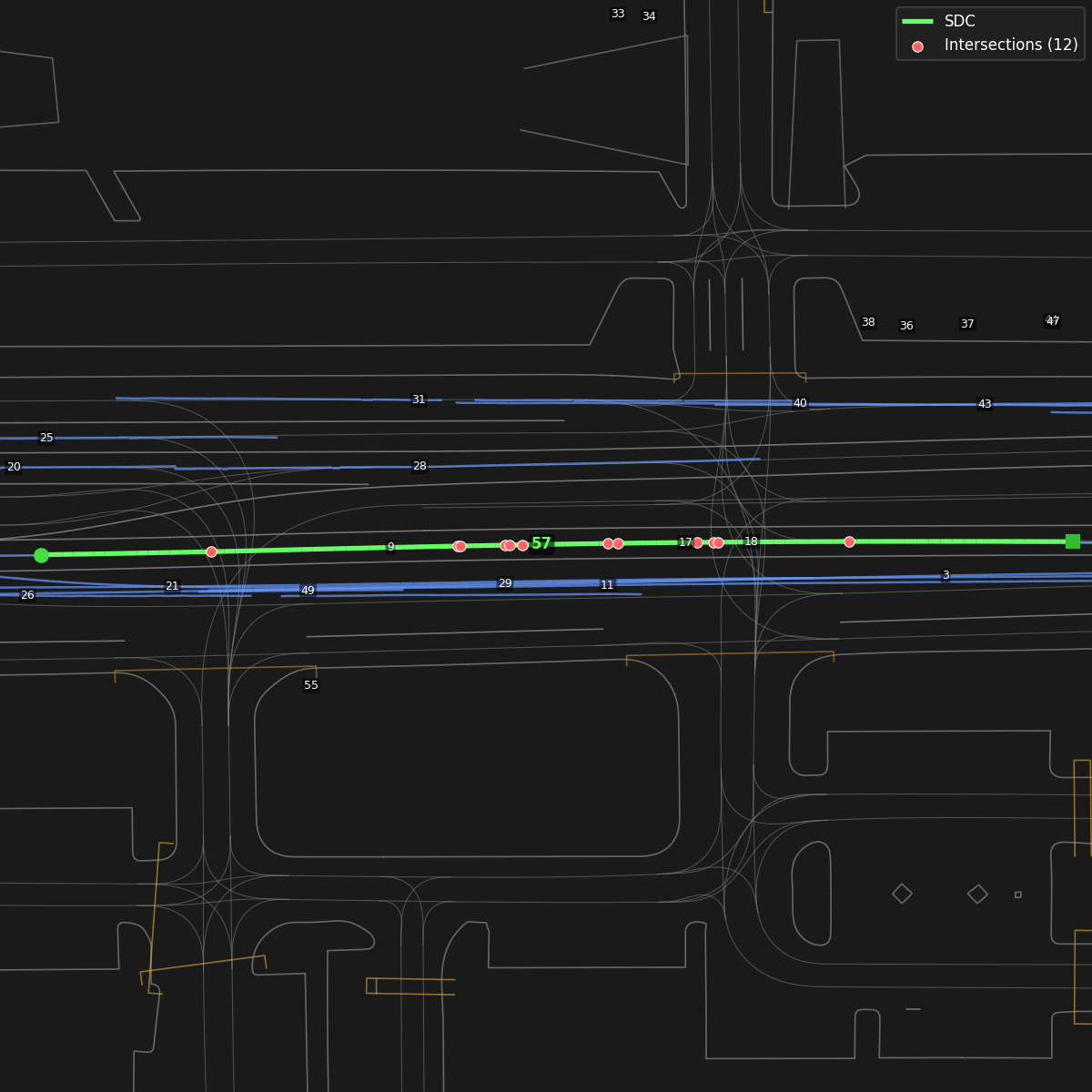}\\[4pt]
    \includegraphics[width=0.32\linewidth]{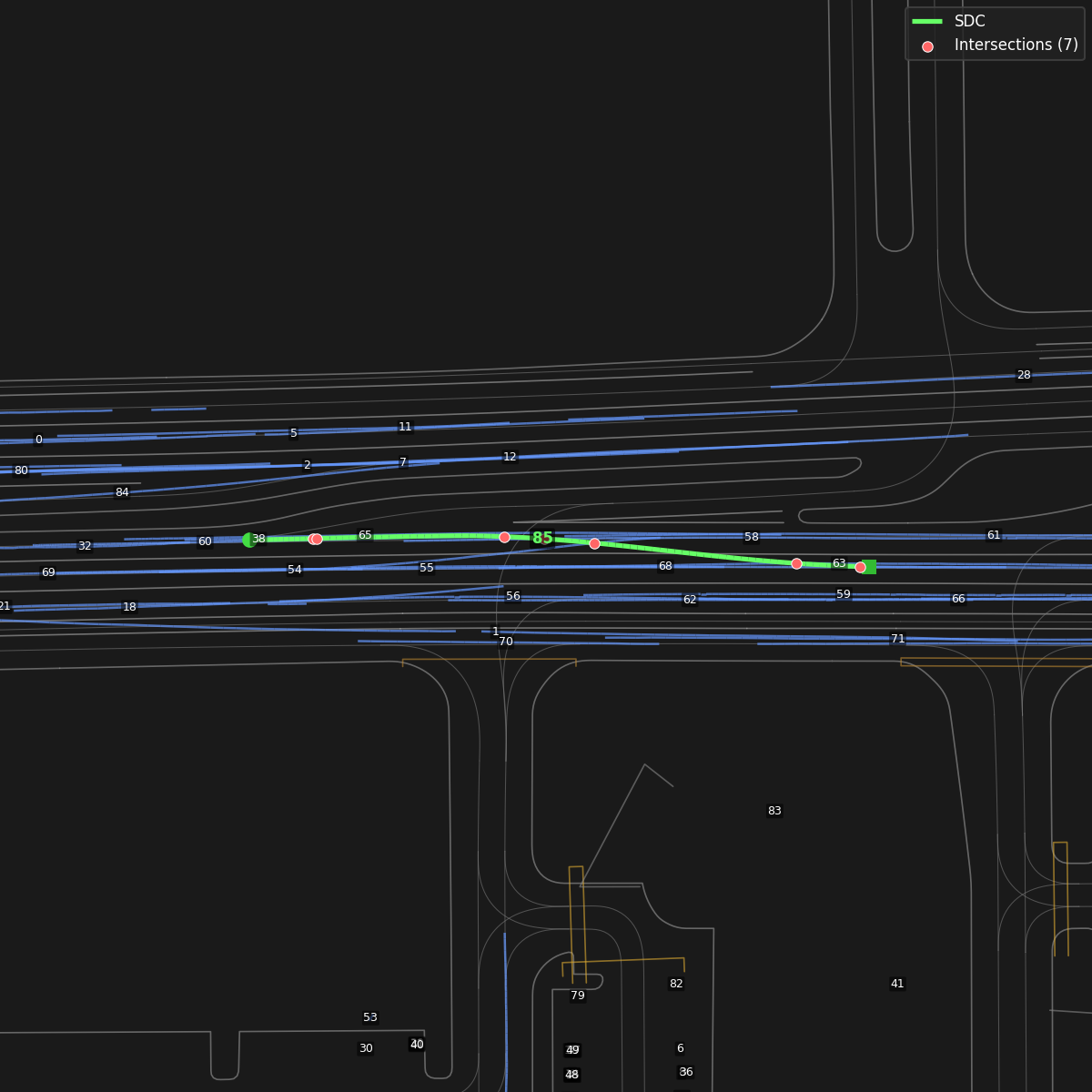}%
    \hfill
    \includegraphics[width=0.32\linewidth]{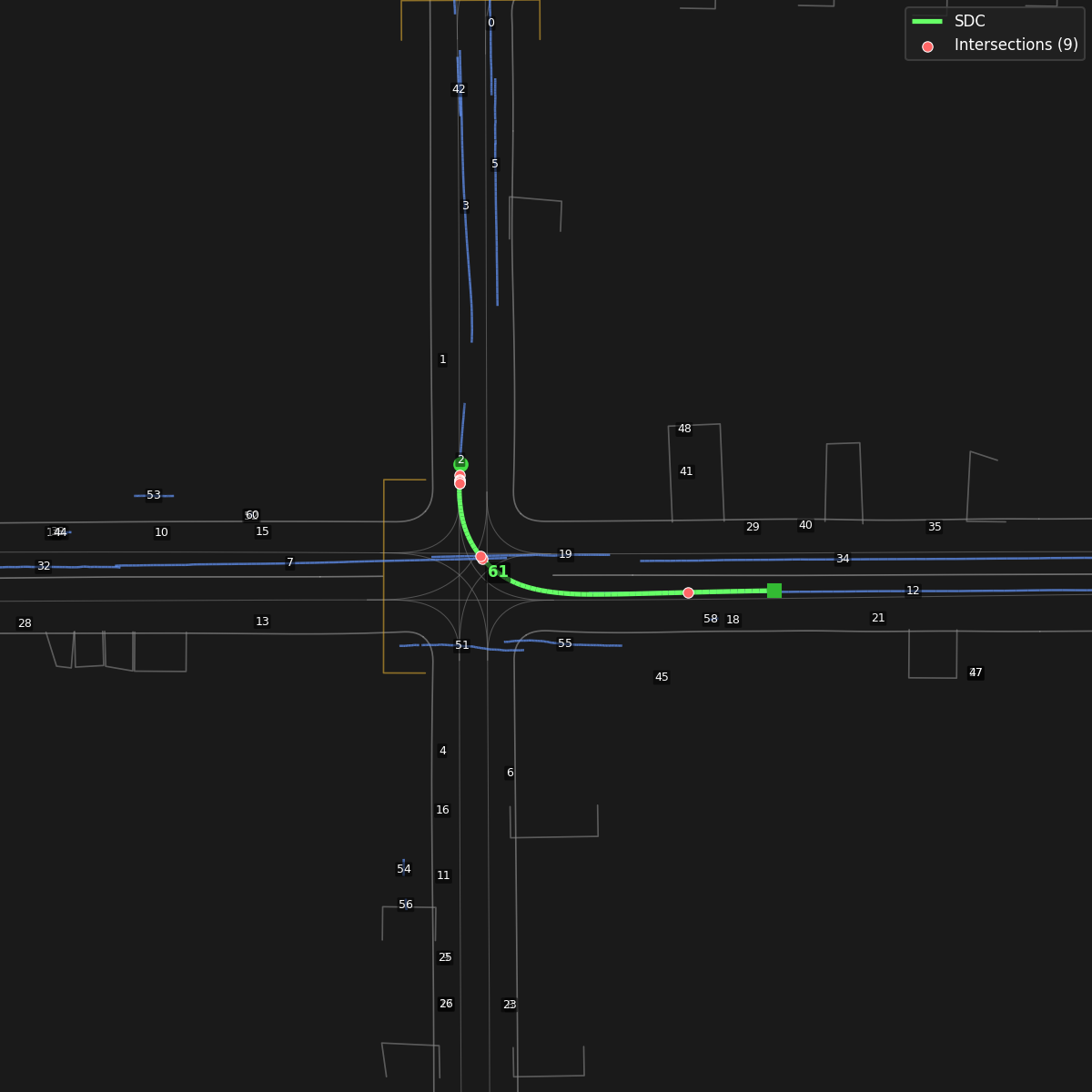}%
    \hfill
    \includegraphics[width=0.32\linewidth]{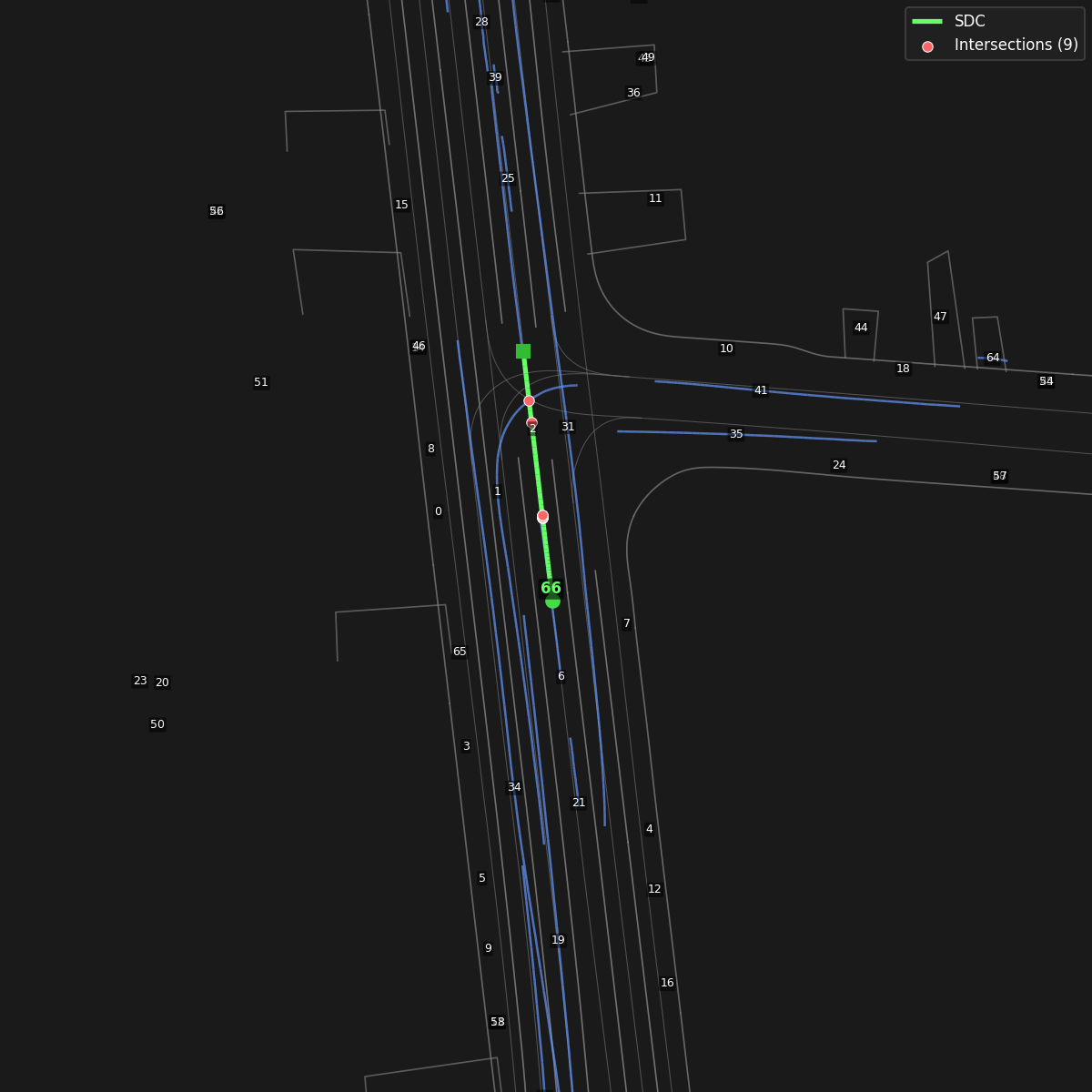}\\[4pt]
    \includegraphics[width=0.32\linewidth]{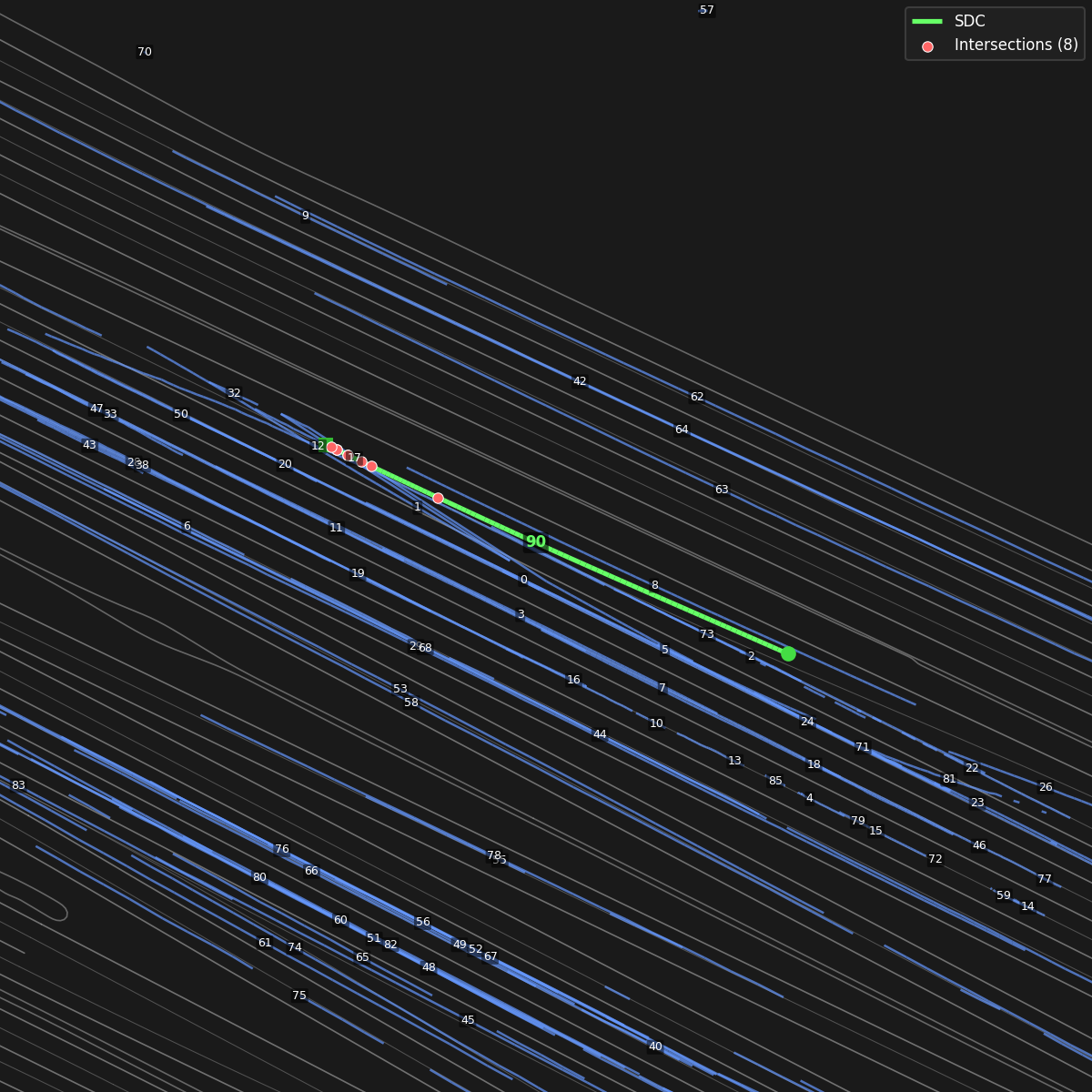}%
    \hfill
    \includegraphics[width=0.32\linewidth]{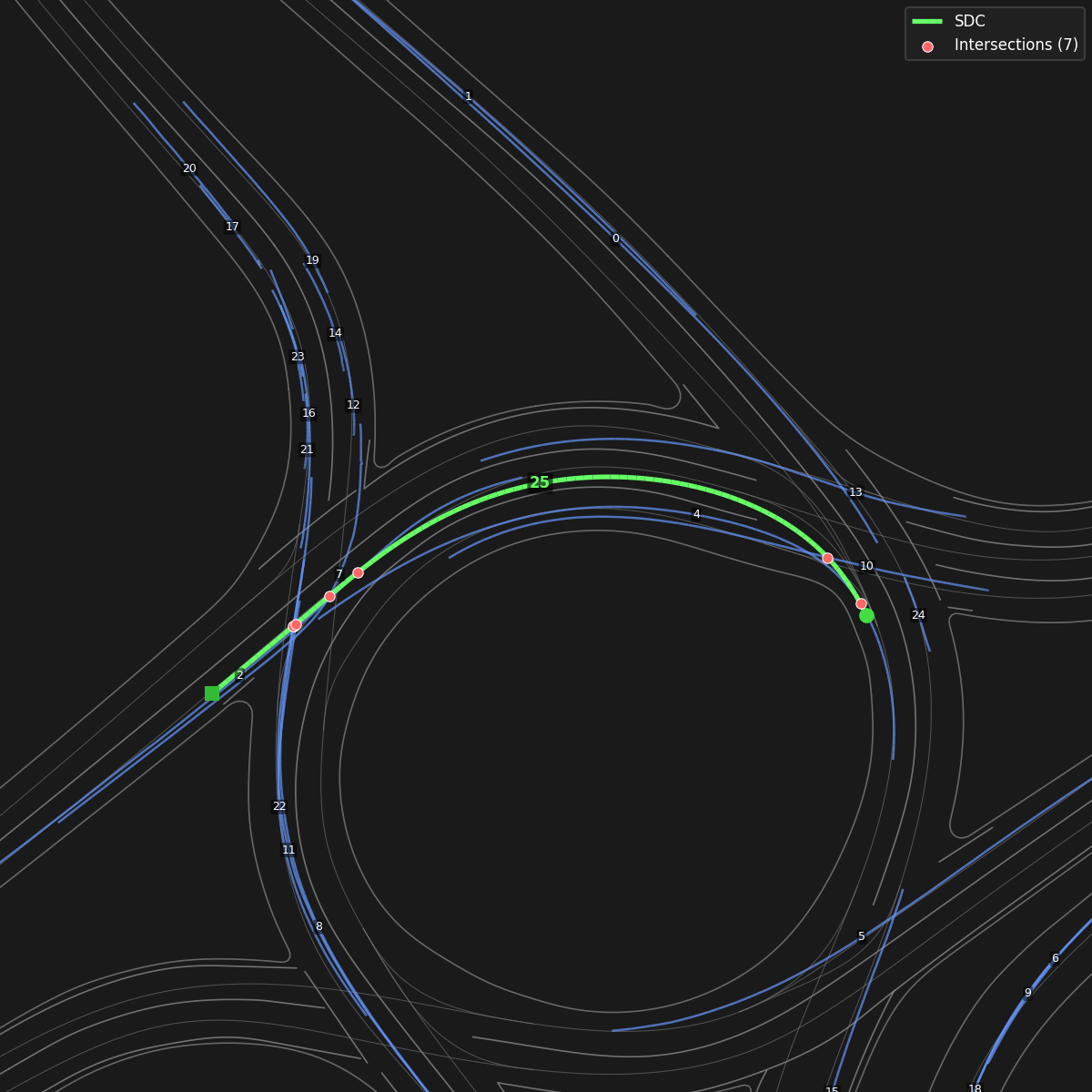}%
    \hfill
    \includegraphics[width=0.32\linewidth]{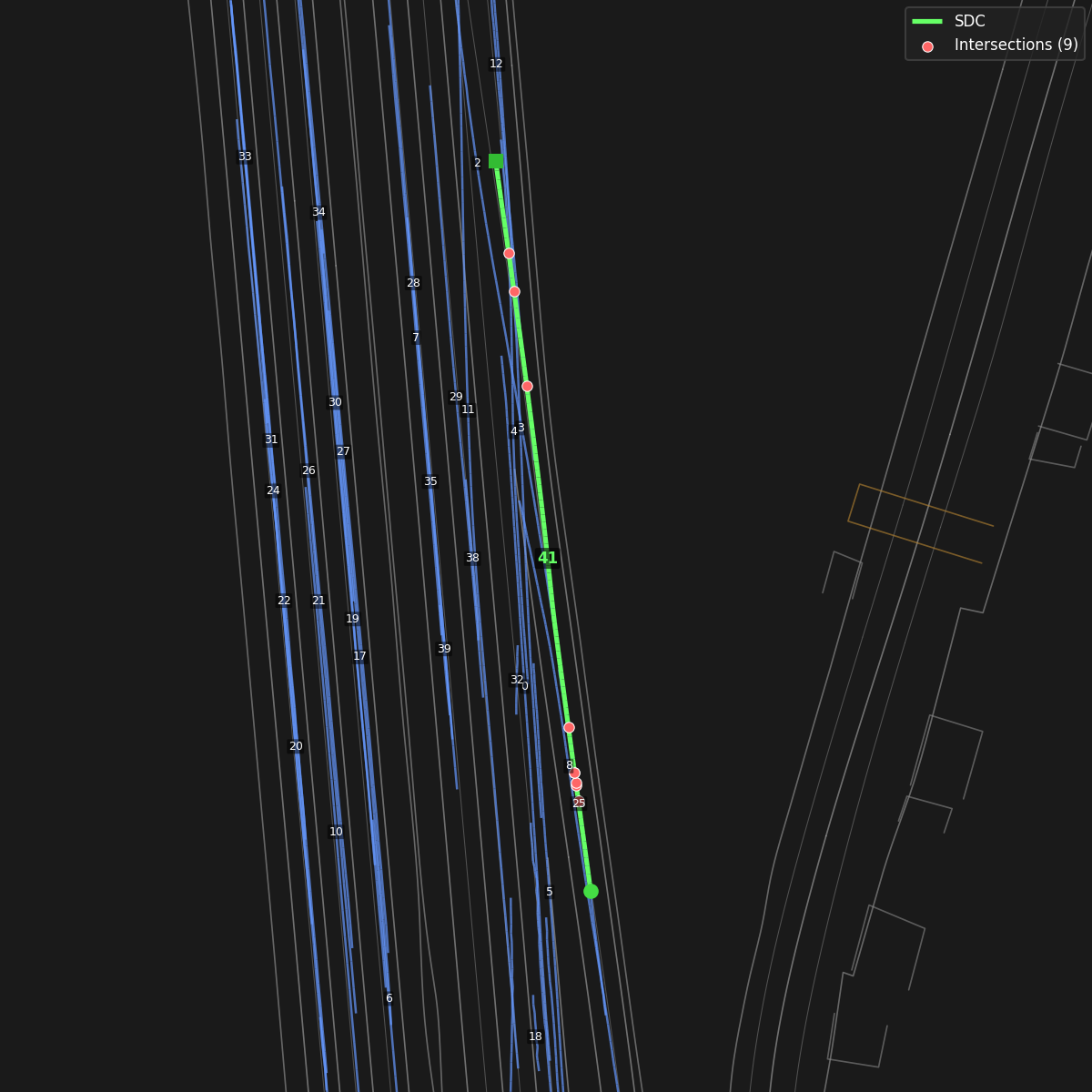}
    \caption{Nine example scenarios from the selected interactive subset. The SDC trajectory is shown in green, other agents in blue, and trajectory intersection points with other logs in red.}
    \label{fig:interactive_scenario_examples}
\end{figure}

We score each scenario by counting the number of segment-level intersections between the SDC trajectory and all other agent trajectories, optionally filtering crossings that meet a minimum acute-angle threshold (to exclude near-parallel overlaps, such as lane changes). From a pool of 10{,}000 held-out validation scenarios, we rank by intersection count and select the top 200 most interactive scenes. Figure~\ref{fig:interactivity_distribution} shows the resulting intersection count distributions across the full dataset and the selected subset, and Figure~\ref{fig:interactive_scenario_examples} shows nine representative examples from the selected set.

\subsection{Metrics}
\label{sec:eval_metrics}

We report the following metrics across all experiments. Unless noted, all metrics are computed per active (i.e., controlled) agent per episode and averaged across agents and scenarios.

\paragraph{Score.}
An agent scores 1 if it reaches its goal without any collision or off-road event during the episode, and 0 otherwise. It jointly captures all failure modes and is a useful aggregate metric.

\paragraph{Completion rate.}
The fraction of agents that reach their goal position (within $\delta = 2$ meters) before episode end, regardless of whether a collision or off-road event occurred.

\paragraph{Collision rate.}
The fraction of episodes in which the agent is involved in at least one collision with another vehicle.

\paragraph{At-fault collision rate.}
A subset of the collision criteria taken from NAVSIM \citep{dauner2024navsim}. A collision is attributed to an agent if (i) the other vehicle is in front of the agent at the time of impact, and (ii) the agent's velocity vector points toward the other vehicle. This filters out collisions in which the agent was rear-ended or struck laterally by an inattentive partner.

\paragraph{Collision severity ($\Delta v$).}
\label{sec:collision_severity_metric}
Beyond the binary collision indicator, we measure the severity of each at-fault collision event using the change in velocity ($\Delta v$) imparted to the agent at impact. Following the impulse-momentum formulation used in \citep{waymo_safety_impact_2025}, the Delta-V of agent $i$ in a collision with partner $j$ is
\begin{equation}
    \Delta v_i = \frac{m_j}{m_i + m_j}\,(1 + e)\,\bigl(\vec{v}_j - \vec{v}_i\bigr) \cdot \hat{n},
\end{equation}
where $\hat{n}$ is the unit collision normal (taken as the vector from agent $i$'s center to agent $j$'s center at impact), $e = 0.1$ is the coefficient of restitution for vehicle-to-vehicle crashes, and the dot product is clipped at zero to ignore separating velocities. Masses are proxied from bounding-box footprint for vehicles (anchored at $1500\,\mathrm{kg}$ for a $4.5\,\mathrm{m} \times 1.8\,\mathrm{m}$ reference sedan) and fixed for vulnerable road users ($75\,\mathrm{kg}$ for pedestrians, $90\,\mathrm{kg}$ for cyclists). $\Delta v$ is one of the strongest predictors of injury risk in vehicle-to-vehicle crashes \citep{waymo_safety_impact_2025} and lets us distinguish low-impact contacts (e.g.\ parking-lot taps) from high-energy collisions even when the binary collision rate is identical.

\paragraph{Off-road rate.}
The fraction of episodes in which the agent crosses a road edge boundary, detected by checking for intersection between the agent bounding box and any road edge polyline.

\paragraph{Route progress ratio.}
Following \citep{gulino2023waymax}, we measure how far along its expert reference trajectory each agent travels. At each timestep $t$, we find the closest point $x(t)$ on the agent's logged trajectory and compute its arc-length distance $d_{x(t)}$ from the start of the path. The route progress ratio is
\begin{equation}
    \rho = \frac{d_{x(t)} - d_p}{d_q - d_p},
\end{equation}
where $d_p$ and $d_q$ are the arc-length distances to the initial and final positions of the logged trajectory, respectively. A value of $\rho = 1$ means the agent reached its destination; $\rho > 1$ is possible if the agent overshoots. For agents that reach their goal under \textsc{goal\_remove} termination, we set $\rho = 1$ directly, since their position is invalidated upon removal. For all other agents, $\rho$ is computed from the agent's final position at episode end.

\paragraph{Lateral deviation.}
At each timestep $t$ for which the agent is alive, we find the nearest valid point on the agent's expert reference trajectory,
\begin{equation}
    k^*(t) = \arg\min_k \|p_t - q_k\|_2,
\end{equation}
where $p_t$ is the agent's simulated position and $q_k$ is the expert position at reference index $k$. The lateral deviation is
\begin{equation}
    \ell_t = \|p_t - q_{k^*(t)}\|_2.
\end{equation}
We report the mean of $\ell_t$ over alive timesteps. This metric is geometry-aligned rather than time-aligned: it measures cross-track drift from the reference path, independent of whether the agent is early or late along that path.

\paragraph{Longitudinal deviation.}
We also decompose path-following error along the expert trajectory. Let $d_k$ denote the cumulative arc length of the expert trajectory up to reference index $k$. At timestep $t$, using the same nearest reference point $k^*(t)$ as above, the signed longitudinal deviation is
\begin{equation}
    r_t = d_{k^*(t)} - d_t.
\end{equation}
Positive values indicate that the agent is ahead of the time-aligned expert along the route, while negative values indicate that it is behind. We report the mean absolute longitudinal deviation, $\mathbb{E}_t[|r_t|]$, over alive timesteps. Like lateral deviation, this metric is route-aligned rather than strictly time-aligned.

\paragraph{Average displacement error (ADE).}
Finally, we report the standard time-aligned displacement error. At each timestep $t$, we compare the agent's simulated position directly to the expert position at the same timestep:
\begin{equation}
    \mathrm{ADE}_t = \|p_t - q_t\|_2.
\end{equation}
We average this quantity over all alive timesteps with a valid expert reference state. Unlike the lateral and longitudinal deviations, ADE is strictly time-aligned and therefore penalizes both spatial deviation and timing error.

\section{Mapping Agent Experience to Human Time}
\label{sec:exp_to_time_calc}
 
We train self-play RL agents on 20 billion transitions. Since Waymo scenarios are discretized at 10\,Hz, each transition $(o_t, a_t)$ corresponds to 0.1 seconds of real time, placing the total training experience at approximately 63 years of driving.
 
For comparison, SMART \citep{wu2024smart} was trained on the full Waymo Open Motion Dataset, which contains 500,000 training scenarios. Each scenario contributes roughly 90 transitions of SDC trajectory data at 10\,Hz, amounting to approximately 45 million transitions in total. The open-sourced SMART-CLSFT checkpoint was trained on all agents in each scene rather than the SDC alone; assuming an average of 5 agents per scenario, this corresponds to roughly 225 million transitions.
 
Our own checkpoints are trained on subsets of 67, 200, 1,200, and 12,000 maps, each contributing approximately 90 transitions per scenario.

2,500 x claim in the abstract comes from 200 scenarios $\times$ 9 seconds each $=$ 30 minutes. 500,000 $\times$ 9 seconds = 75,000 minutes. 30 minutes / 75,000 minutes = 0.0004.

\FloatBarrier
\section{Additional Results}

\subsection{The Role of Scenario Metadata}
\label{sec:role_of_meta_data}

\paragraph{Scenario diversity is essential for learning general policies.} 
Aside from human driving data, a cheaper source of simulation grounding data is scenario \textit{metadata}: road graphs, initial positions, and velocities. Recent work has shown that regularized self-play RL grounded by target-city metadata can adapt driving policies to new cities~\citep{wang2026nomad}. A natural follow-up question is how much the diversity provided by metadata matters for training generalizable policies, which is what we explore here. We train regularized and unregularized self-play RL agents on subsets $\mathcal{M}_k$ with $|\mathcal{M}_k| \in \{10, 100, 1{,}000, 10{,}000, 50{,}000\}$ scenarios, holding the BC anchors $\tau^n$ and reward function $r$ fixed. This isolates the effect of environment initialization and diversity besides the agent behaviors.

\begin{figure}[ht]
    \centering
    \includegraphics[width=1\linewidth]{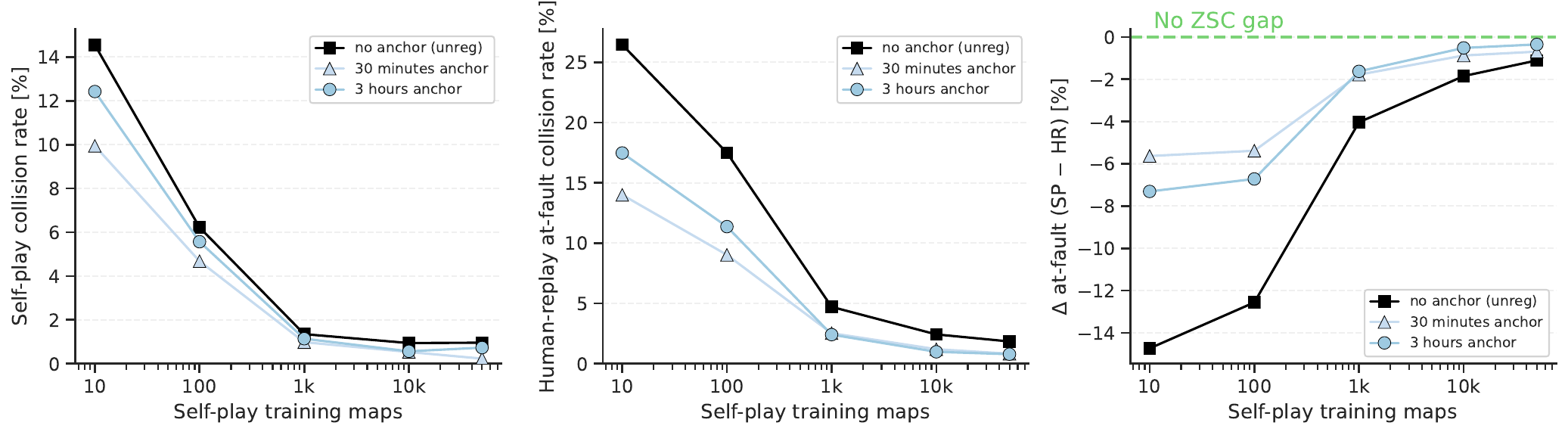}
    \caption{\textbf{Scaling scenario metadata}. The unregularized self-play baseline is shown in black; shades of blue correspond to regularized policies trained with different BC anchors, with darker shades indicating more anchor data. \textit{Left:} collision rate in self-play, where all agents are controlled by the same policy on a held-out validation set. \textit{Center:} at-fault collision rate, the fraction of collisions caused by the controlled agent (See cartoon in Figure \ref{fig:cartoon_to_explain_evals}). \textit{Right:} Gap between self-play and human-replay performance (here referred to as zero-shot coordination; $\Delta_\mathrm{ZSC}$). Concretely, it is difference in the at-fault collision rate between the self-play and human-replay settings.}
    \label{fig:reg_self_play_data_requirements}
\end{figure}

We find that the number of training scenarios (a proxy for map diversity) is an important ingredient for generalization, both to held-out maps and to the human-replay population. As shown in Figure \ref{fig:reg_self_play_data_requirements}, both unregularized and regularized self-play improve drastically with the amount of metadata. For unregularized self-play, the at-fault collision rate drops from 14\% at 10 scenarios to 0.5-1\% at 50k scenarios, and the human-replay collision rate falls from 25.2\% to 2\% over the same range. Regularized self-play follows the same trend and reaches lower absolute values: with a 30-min BC anchor, the human-replay at-fault collision rate drops from 14\% at 10 scenarios to 0.7\% at 50k scenarios. The gap between the self-play performance (pairing policy with itself) and the human-replay population approaches ~0.2\% for regularized policies, and is 1.5\% for unregularized self-play.

\subsection{Human driving data}

\begin{figure}[ht]
    \centering
    \includegraphics[width=0.95\linewidth]{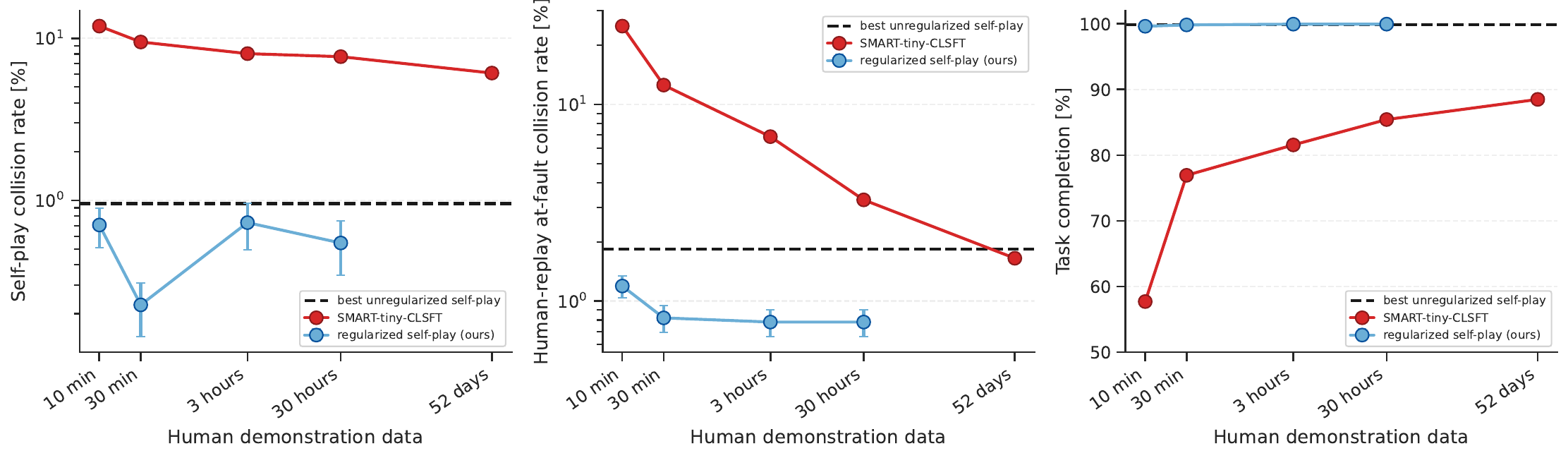}
    \caption{Scaling human driving data for reg. self-play RL; Same as Figure \ref{fig:human_data_requirements} but with the collision rates on a log scale.}
    \label{fig:human_data_reqs_semilog}
\end{figure}

 

\subsection{Regularization keeps RL policies close to human anchors}
\label{sec:solution_space}

\begin{figure}[ht]
\centering
\includegraphics[width=1\linewidth]{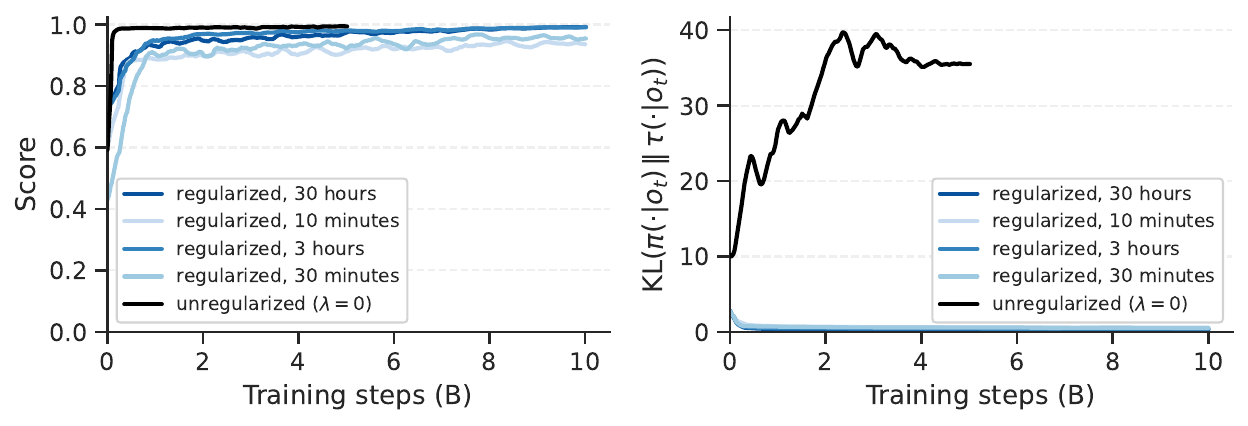}
\caption{\textbf{Regularized self-play remains close to the anchor distribution while unregularized self-play diverges.} Both agents converge to effective driving strategies (left), but their action distributions differ, as measured by KL divergence between observation-conditioned action distributions (right). Regularized policies stay near the anchor; unregularized policies diverge monotonically.}
\label{fig:learning_curves_diff}
\end{figure}

Figure~\ref{fig:learning_curves_diff} shows task completion and KL divergence to the anchor policy over training. Both regularized and unregularized agents converge to comparably effective strategies in terms of goal completion and collision avoidance, yet the underlying action distributions diverge substantially. Without regularization, the agent drifts freely through the space of competent policies, converging far from human behavior; KL divergence increases monotonically throughout training. Regularization constrains the trajectory through policy space without restricting the set of achievable outcomes: the agent remains free to discover effective strategies, but the penalty keeps those strategies within the behavioral distribution of human driving. The result is an agent that is both capable and closer to the distribution of human driving.

\FloatBarrier
\subsection{Distributional Realism: Waymo Open Sim Agent Challenge}
\label{sec:wosac_full}

\begin{figure}[ht]
    \centering
    \includegraphics[width=1\linewidth]{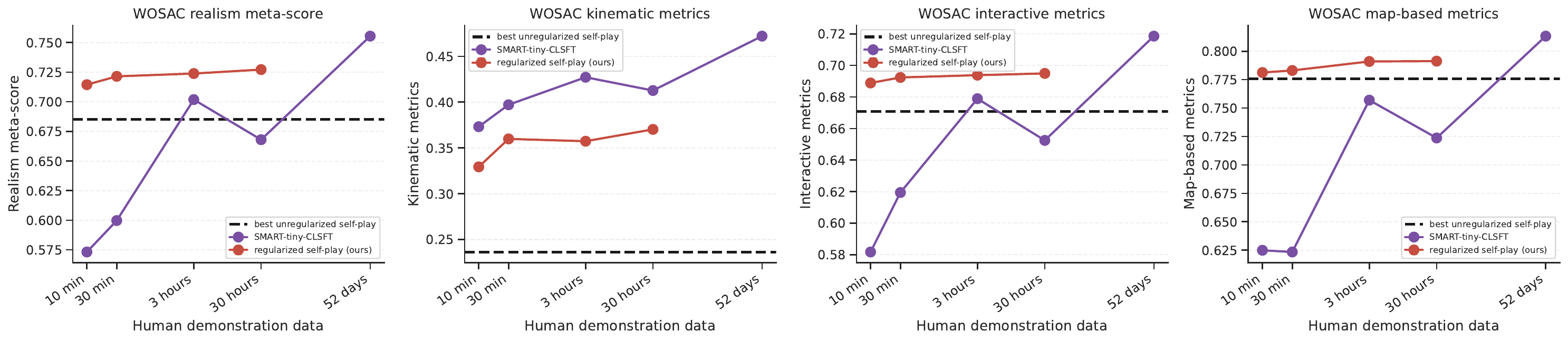}
    \caption{WOSAC meta-scores and group metrics.}
    \label{fig:wosac_meta}
\end{figure}

\begin{figure}[ht]
    \centering
    \includegraphics[width=1\linewidth]{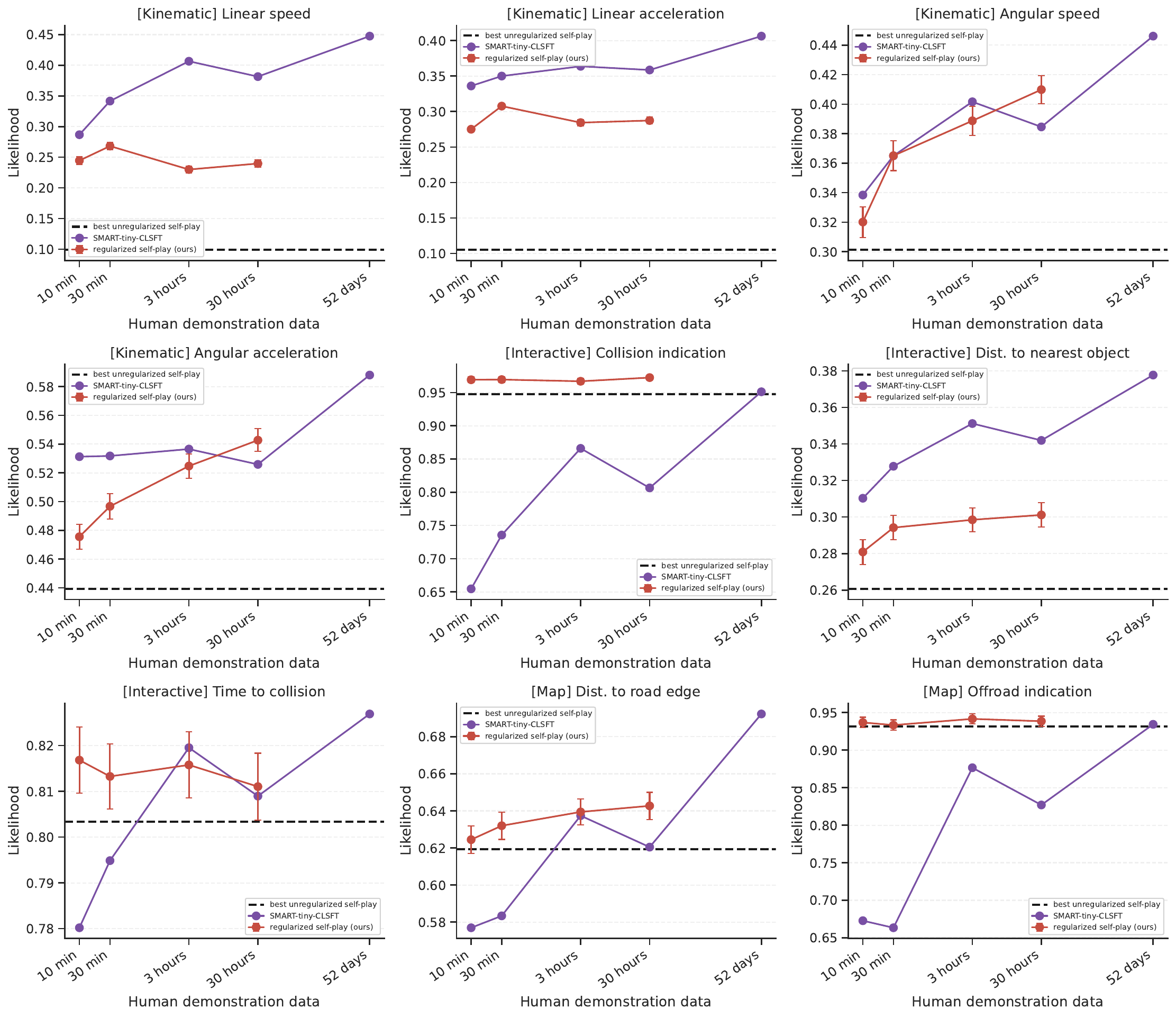}
    \caption{WOSAC submetrics}
    \label{fig:wosac_sub}
\end{figure}

Figure~\ref{fig:wosac_meta} reports the WOSAC \citep{montali2023waymo} realism meta-score alongside its three group
metrics (kinematic, interactive, and map-based); Figure~\ref{fig:wosac_sub} breaks down all nine submetrics that together make up the meta-score.

\paragraph{Unregularized self-play.}
Unregularized self-play achieves a WOSAC meta-score of 0.68, with the largest deficits in the kinematic (0.22) and interactive groups. As shown in Figure~\ref{fig:wosac_sub}, these policies produce low likelihoods particularly in linear speed, acceleration, and distance to nearest object.

\paragraph{Regularized self-play.}
Adding regularization improves the meta-score to 0.725, with gains over unregularized self-play across every metric. The score is largely insensitive to additional data.

\paragraph{SMART-tiny CLSFT.}
SMART trained on 52 days of human data achieves the highest meta-score of 0.755, despite a worse collision rate and task completion across all data bins (Table~\ref{tab:human_data_results}). This result is consistent with the SMART-tiny CLSFT results reported on the CATK github repository.

\FloatBarrier
\subsection{Safety analysis}

\begin{table}[ht]
\centering
\caption{Comparing unregularized and regularized self-play policies on 10k random validation split. Long.\ L2 and Lat.\ L2 are the displacement errors from the human trajectory decomposed along the direction of travel and perpendicular to it, and ADE is the average displacement error over the episode time-aligned to the logs (all in meters). Lower is better throughout. Best value per column in \textbf{bold}.}
\label{tab:main_comparison}
\begin{tabular}{l|rrrr}
\toprule
 & \multicolumn{4}{c}{Human-replay (interactive)} \\
\makecell{Method} & At-fault (\%) $\downarrow$ & Long. L2 $\downarrow$ & Lat. L2 $\downarrow$ & Time-aligned ADE $\downarrow$ \\
\midrule
Unregularized & $2.1 \pm 0.5$ & $13.327 \pm 0.129$ & $2.390 \pm 0.148$ & $14.074 \pm 0.182$ \\
Regularized (ours) & $\bm{0.7 \pm 0.3}$ & $\bm{5.559 \pm 0.077}$ & $\bm{1.274 \pm 0.029}$ & $\bm{5.927 \pm 0.076}$ \\
\bottomrule
\end{tabular}
\end{table}

\FloatBarrier
\subsection{Single and multi-agent RL}
\label{sec:single_vs_multi_agent}

\begin{figure}[ht]
    \centering
    \includegraphics[width=1\linewidth]{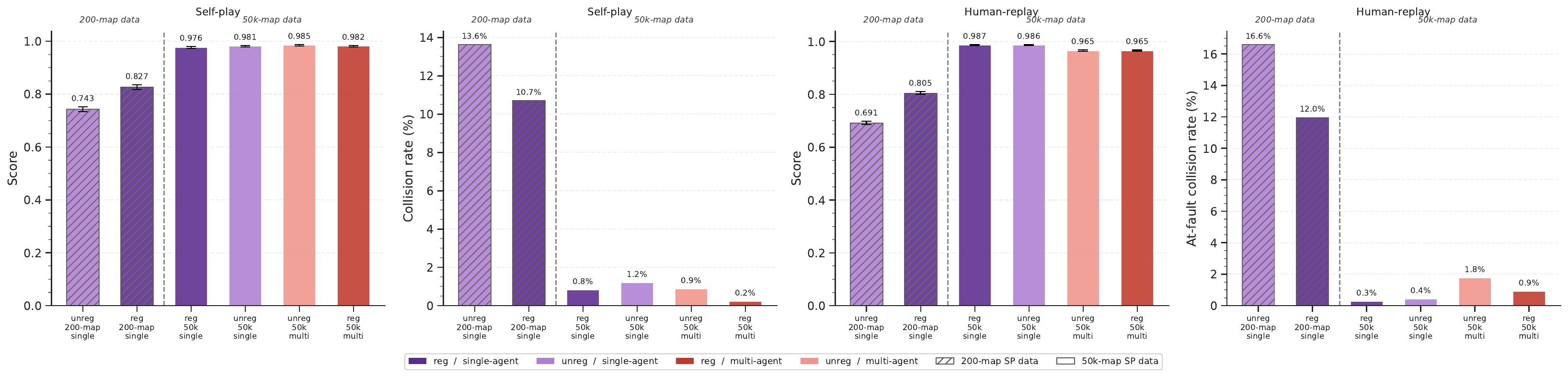}
    \caption{Single vs. multi-agent experiments. Purple bar plots represent performance of policies trained in a single-agent setting; Red barplots are policies trained in a multi-agent (self-play) setting.}
    \label{fig:single_multi_agent}
\end{figure}

\FloatBarrier
\section{Extended limitations}
\label{sec:extended_limitations}

\paragraph{Failure modes and directions for improvement.}
We perform an additional analysis to better understand the limitations of the resulting regularized policies. 
To improve the signal of the analysis, we evaluate on a curated set of \textit{interactive} scenarios within the held-out set, that is, filter for scenarios that contain dense multi-agent interactions such as merges, unprotected turns, and yielding (details in Appendix \ref{sec:eval_interactive_filtering}).

Table~\ref{tab:interactive_comparison_idm_hr} shows that (at-fault) collision rates increase noticeably in these interactive scenarios, even for the best regularized policy (2.1-2.8\%) and SMART-tiny-CLSFT trained on 52 days of data (2.7\%). We also share several representative failure modes on the webpage \url{https://spiced-self-play.com/} (see failure modes).

A likely reason for the increased collision rates for the self-play policies is that the Waymo scenarios that we train in during self-play are small (since they are constructed from a 9-second log), and agent interactions are relatively sparse (see Figure \ref{fig:interactivity_distribution} for the distribution of intersections between agent logs), so the RL agent only occasionally trains on transitions that improve difficult coordination situations. 

We outline several directions for future work that could improve robustness:
\begin{enumerate}[noitemsep, topsep=0pt]
    \item \textbf{Curriculum learning based on advantage.} Each scenario can be treated as a level whose difficulty is measured by the agent's average advantage. Upsampling scenarios proportionally to their advantage would concentrate training signal on cases the agent finds difficult, naturally increasing exposure to rare but safety-critical situations such as sudden cut-ins and stationary obstacles.
    \item \textbf{Domain randomization.} Masking out the observation of a ratio of agents within each scenario ("blind" agents \cite{cusumano2025robust}) and adding noise to the dynamics or partner features provides a targeted form of domain randomization that could make policy behavior more cautious.
    \item \textbf{Adversarial fine-tuning.} A third training stage that fine-tunes on a curated set of adversarial human data would expose the policy to scenarios where the other agents in the scene do not respond to it.
    \item \textbf{Human-like opponents.} Occasionally replacing the self-play opponent with the BC anchor rather than a copy of the RL policy would expose the agent to more human-like partner behavior throughout training.
    \item \textbf{Stronger anchor policy.} The BC anchor is itself a limiting factor: our best anchor achieves a closed-loop score of 0.66 (Table~\ref{tab:anchor_results}), and a stronger anchor, whether through architectural improvements or additional data, would give the KL regularizer a more reliable behavioral target. 
\end{enumerate}

\definecolor{tierbest}{HTML}{6FCF6A}
\definecolor{tiersecond}{HTML}{DFF04B}
\definecolor{tierthird}{HTML}{FBF4D0}
\definecolor{tierunregbest}{HTML}{D9D9D9}
\begin{table}[ht]
\centering
\caption{Interactive evaluation across all scaling checkpoints. All metrics are computed on the interactive validation subset; policies are rolled out in each of the 200 scenarios 10 times. Top-3 values per column are highlighted (\colorbox{tierbest}{best}, \colorbox{tiersecond}{2nd}, \colorbox{tierthird}{3rd}); best value additionally in bold. \colorbox{tierunregbest}{Gray} marks the best unregularized self-play value per column. IDM results are not available for SMART (indicated by ---).}
\label{tab:interactive_comparison_idm_hr}
\resizebox{\textwidth}{!}{%
\begin{tabular}{rr|rr|rrrr}
\toprule
 & & \multicolumn{2}{c|}{Score} & \multicolumn{4}{c}{Collision rates} \\
\makecell{Self-play maps \\ (metadata)} & \makecell{Anchor data \\ (human demos)} & HR Score $\uparrow$ & IDM Score $\uparrow$ & IDM At-fault (\%) $\downarrow$ & HR At-fault (\%) $\downarrow$ & IDM Coll. (\%) $\downarrow$ & HR Coll. (\%) $\downarrow$ \\
\midrule
10 & 0 (unreg.) & $0.312 \pm 0.010$ & $0.296 \pm 0.010$ & $42.8 \pm 1.1$ & $46.2 \pm 1.1$ & $46.6 \pm 1.1$ & $50.1 \pm 1.1$ \\
100 & 0 (unreg.) & $0.598 \pm 0.011$ & $0.577 \pm 0.011$ & $28.9 \pm 1.0$ & $29.9 \pm 1.0$ & $34.6 \pm 1.1$ & $34.3 \pm 1.0$ \\
1k & 0 (unreg.) & $0.868 \pm 0.007$ & $0.842 \pm 0.008$ & $5.8 \pm 0.5$ & $7.6 \pm 0.6$ & $10.1 \pm 0.7$ & $12.2 \pm 0.7$ \\
10k & 0 (unreg.) & $0.891 \pm 0.007$ & $0.876 \pm 0.007$ & \cellcolor{tierunregbest} $3.2 \pm 0.4$ & \cellcolor{tierunregbest} $4.1 \pm 0.4$ & $9.0 \pm 0.6$ & $10.2 \pm 0.7$ \\
50k & 0 (unreg.) & \cellcolor{tierunregbest} $0.908 \pm 0.006$ & \cellcolor{tierunregbest} $0.893 \pm 0.007$ & $3.8 \pm 0.4$ & $4.9 \pm 0.5$ & \cellcolor{tierunregbest} $7.6 \pm 0.6$ & \cellcolor{tierunregbest} $8.7 \pm 0.6$ \\
\midrule
10 & 30 minutes & $0.425 \pm 0.011$ & $0.432 \pm 0.011$ & $33.1 \pm 1.0$ & $34.6 \pm 1.1$ & $36.6 \pm 1.1$ & $37.6 \pm 1.1$ \\
10 & 3 hours & $0.361 \pm 0.011$ & $0.371 \pm 0.011$ & $37.3 \pm 1.1$ & $39.6 \pm 1.1$ & $39.8 \pm 1.1$ & $43.2 \pm 1.1$ \\
100 & 30 minutes & $0.722 \pm 0.010$ & $0.661 \pm 0.010$ & $16.8 \pm 0.8$ & $18.0 \pm 0.8$ & $22.4 \pm 0.9$ & $23.6 \pm 0.9$ \\
100 & 3 hours & $0.658 \pm 0.010$ & $0.629 \pm 0.011$ & $21.8 \pm 0.9$ & $24.0 \pm 0.9$ & $25.5 \pm 1.0$ & $28.2 \pm 1.0$ \\
1k & 30 minutes & $0.897 \pm 0.007$ & $0.858 \pm 0.008$ & $4.4 \pm 0.5$ & $5.9 \pm 0.5$ & $8.4 \pm 0.6$ & $9.2 \pm 0.6$ \\
1k & 3 hours & $0.886 \pm 0.007$ & $0.866 \pm 0.008$ & $5.3 \pm 0.5$ & $7.0 \pm 0.6$ & $9.3 \pm 0.6$ & $10.2 \pm 0.7$ \\
10k & 10 minutes & $0.916 \pm 0.006$ & $0.858 \pm 0.008$ & $3.1 \pm 0.4$ & $3.0 \pm 0.4$ & $8.3 \pm 0.6$ & $6.8 \pm 0.6$ \\
10k & 30 minutes & $0.926 \pm 0.006$ & $0.892 \pm 0.007$ & $3.5 \pm 0.4$ & \cellcolor{tiersecond} $2.4 \pm 0.3$ & $7.9 \pm 0.6$ & $7.1 \pm 0.6$ \\
10k & 3 hours & $0.906 \pm 0.006$ & $0.873 \pm 0.007$ & $3.0 \pm 0.4$ & $3.5 \pm 0.4$ & $7.7 \pm 0.6$ & $7.9 \pm 0.6$ \\
10k & 30 hours & $0.925 \pm 0.006$ & \cellcolor{tiersecond} $0.904 \pm 0.007$ & \cellcolor{tiersecond} $2.6 \pm 0.4$ & $3.5 \pm 0.4$ & \cellcolor{tierthird} $5.9 \pm 0.5$ & $6.0 \pm 0.5$ \\
50k & 10 minutes & $0.923 \pm 0.006$ & $0.883 \pm 0.007$ & $3.1 \pm 0.4$ & $3.0 \pm 0.4$ & $7.4 \pm 0.6$ & $6.9 \pm 0.6$ \\
50k & 30 minutes & \cellcolor{tierthird} $0.931 \pm 0.006$ & $0.890 \pm 0.007$ & \cellcolor{tierthird} $2.8 \pm 0.4$ & \cellcolor{tierthird} $2.6 \pm 0.4$ & \cellcolor{tiersecond} $5.6 \pm 0.5$ & $6.0 \pm 0.5$ \\
50k & 3 hours & \cellcolor{tiersecond} $0.935 \pm 0.005$ & $0.890 \pm 0.007$ & $3.6 \pm 0.4$ & $2.8 \pm 0.4$ & $6.5 \pm 0.5$ & \cellcolor{tiersecond} $5.2 \pm 0.5$ \\
50k & 30 hours & \cellcolor{tierbest} $\bm{0.949 \pm 0.005}$ & \cellcolor{tierbest} $\bm{0.908 \pm 0.006}$ & \cellcolor{tierbest} $\bm{2.2 \pm 0.3}$ & \cellcolor{tierbest} $\bm{2.1 \pm 0.3}$ & \cellcolor{tierbest} $\bm{5.2 \pm 0.5}$ & \cellcolor{tierbest} $\bm{4.2 \pm 0.4}$ \\
\midrule
--- & 10 min (SMART) & $0.048 \pm 0.005$ & --- & --- & $35.0 \pm 1.1$ & --- & $43.9 \pm 1.1$ \\
--- & 30 min (SMART) & $0.148 \pm 0.008$ & --- & --- & $24.5 \pm 1.0$ & --- & $30.9 \pm 1.0$ \\
--- & 3 hours (SMART) & $0.319 \pm 0.010$ & --- & --- & $15.3 \pm 0.8$ & --- & $21.2 \pm 0.9$ \\
--- & 30 hours (SMART) & $0.376 \pm 0.011$ & --- & --- & $6.4 \pm 0.5$ & --- & $11.6 \pm 0.7$ \\
--- & 52 days BC (SMART) & $0.383 \pm 0.011$ & --- & --- & $4.5 \pm 0.5$ & --- & $7.9 \pm 0.6$ \\
--- & 52 days CLSFT (SMART) & $0.433 \pm 0.011$ & --- & --- & $2.7 \pm 0.4$ & --- & \cellcolor{tierthird} $5.4 \pm 0.5$ \\

\bottomrule
\end{tabular}}
\end{table}

\FloatBarrier
\subsection{SMART model performance with and without finetuning}
\label{sec:smart_model_full_res}

The 52-day IL baseline results in Table~\ref{tab:human_data_results} are obtained from the CAT-K fine-tuned SMART model trained on the full 500k-scenario Waymo training set \citep{zhang2025closed}, which achieves the strongest imitation-learning performance (training details in Appendix \ref{sec:smart_model_training_details}). For completeness, Table~\ref{tab:smart_finetuning_results} reports both raw and fine-tuned SMART checkpoints trained on subsets of the Waymo dataset. Although fine-tuning generally improves route completion, the pre-finetuning SMART checkpoints consistently achieve lower collision and off-road rates. We therefore report the raw checkpoints in the main paper, as they yield the strongest overall baseline performance.

\begin{table}[ht]
\centering
\caption{SMART performance with and without CATK \citep{zhang2025closed} fine-tuning on 10,000 held-out validation scenarios. The main paper reports the strongest-performing variant at each data scale; for SMART, these correspond to the non-fine-tuned checkpoints shown here. Fine-tuned rows denote the same checkpoints after closed-loop supervised fine-tuning.}
\label{tab:smart_finetuning_results}
\vspace{-0.5em}
\resizebox{\textwidth}{!}{%
\begin{tabular}{ll|rrr|rrrrr}
\toprule
 & & \multicolumn{3}{c|}{Self-play / all-agents (test)} & \multicolumn{5}{c}{Human-replay / SDC-only (test)} \\
\makecell{Human demos \\ used} & \makecell{Fine-\\tuned} & Coll. (\%) $\downarrow$ & Off-road (\%) $\downarrow$ & Route prog. (\%) $\uparrow$ & Score $\uparrow$ & Coll. (\%) $\downarrow$ & At-fault (\%) $\downarrow$ & Off-road (\%) $\downarrow$ & Route prog. (\%) $\uparrow$ \\
\midrule
10 min & No  & 11.9 & 55.8 & 84.5 & 0.246 & 32.0 & 25.0 & 18.6 & 57.7 \\
10 min & Yes & 19.2 & 57.7 & 85.1 & 0.216 & 33.3 & 26.9 & 27.3 & 68.5 \\
\midrule
30 min & No  & 9.5 & 55.4 & 85.8 & 0.379 & 17.9 & 12.5 & 16.8 & 76.9 \\
30 min & Yes & 13.4 & 56.9 & 87.0 & 0.311 & 23.3 & 18.3 & 26.4 & 80.4 \\
\midrule
3 hours & No  & 8.0 & 53.6 & 86.2 & 0.518 & 11.4 & 6.9 & 4.5 & 81.5 \\
3 hours & Yes & 10.5 & 54.3 & 87.2 & 0.481 & 14.1 & 10.1 & 8.9 & 85.7 \\
\midrule
30 hours & No  & 7.7 & 53.3 & 86.5 & 0.601 & 6.8 & 3.3 & 1.6 & 85.4 \\
30 hours & Yes & 9.1 & 53.6 & 87.8 & 0.586 & 6.9 & 4.0 & 2.8 & 89.4 \\
\bottomrule
\end{tabular}}
\vspace{-0.8em}
\end{table}

\end{document}